\documentstyle[fullname,lingmacros,tree-dvips]{article}

\makeatletter
\newcounter{EEXAMPLE}
\renewcommand{\theEEXAMPLE}{\alph{EEXAMPLE}}
\newenvironment{AEXAMPLE}{\begin{list}{}
    {\topsep      0pt
     \partopsep   0pt
     \itemsep     .0ex
     
     \usecounter{EEXAMPLE}
     }\vskip-\lastskip}{\end{list}}

\newcommand{\ALPHA}{\mbox{$\alpha$\/}}
\newcommand{\AND}{\mbox{$\wedge$ }}
\newcommand{\BAR}[1]{{\small $\overline{\mbox{#1}}$}}

\newcommand{\BCONDCURRY}[3]{\PRED{#1}(\TERM{#2}\/)(\TERM{#3}\/)}
\newcommand{\BETA}{\mbox{$\beta$}}
\newcommand{\CITE}[1]{\cite{#1}}

\newcommand{\CRT}{\SIGLA{crt}}
\newtheorem{DEFINITION}{Definition}[section]
\newcommand{\DELTA}{\mbox{$\delta$\/}}

\newcommand{\DENOTATION}[1]{\mbox{$[ \! [$#1$] \! ]$}}
\newcommand{\DET}[1]{{\bf #1}}
\newcommand{\DIP}[4]{\begin{tabular}{l}
                     #1\\[1ex]
                     #2 : #3 : #4\\
                     \hline
                     #4
                     \end{tabular}}
\newcommand{\DIPR}[6]{\begin{tabular}{l}
                     #1\\[1ex]
                     #2 : #3 : #4\\
                     \hline
                     #5 : #6
                     \end{tabular}}
\newcommand{\DISCHARGE}{{\bf discharge}}
\newcommand{\DOMAIN}[1]{\mbox{D$_{\mbox{\TYPE{#1}}}$}}
\newcommand{\DRS}{\SIGLA{drs}}

\newcommand{\DRT}{\SIGLA{drt}}

\newcommand{\ENEW}[1]{\refstepcounter{equation}
\label{#1}\item[(\theequation)\hspace{10pt}]}
\newcommand{\ENUMA}[1]{\refstepcounter{equation}
\label{#1}\item[(\theequation)] \begin{AEXAMPLE}}
\newcommand{\ENDENUMA}{\end{AEXAMPLE}}
\newcommand{\EITEM}{\stepcounter{EEXAMPLE}\item[\theEEXAMPLE.]}
\newcommand{\EQUIV}{\mbox{$\equiv$ }}
\newcommand{\ETAL}{{\em et al.\/}}
\newenvironment{EXAMPLE}{\begin{list}{}
    {\topsep      4pt
     \itemsep     .0ex
     \labelwidth  35pt
     \leftmargin  40pt
     
     }}{\end{list}}
\newcommand{\EXISTS}{\mbox{$\exists$}\ }
\newcommand{\EXISTSEXP}[3]{\QEXP{\EXISTS}{#1}{#2}{#3}}
\newcommand{\FOOTNOTE}[1]{}
\newcommand{\FORALL}{\mbox{$\forall$}\ }
\newcommand{\FORALLEXP}[3]{\QEXP{\FORALL}{#1}{#2}{#3}}
\newcommand{\FTYPE}[2]{\mbox{$\langle$}\TYPE{#1},\TYPE{#2}\mbox{$\rangle$}}
\newcommand{\FUNDEF}[1]{\parbox{350pt}{\[\left\{
                        \parbox{300pt}{
                        \begin{description}
                        #1
                        \end{description}
                        }
                        \right.\]}}
\newcommand{\GAMMA}{\mbox{$\gamma$\/}}

\newcommand{\IE}{{i.e.}}
\def\ifempty#1{\@@ifempty #1\@emptymarkA\@emptymarkB}%
\def\@@ifempty#1#2\@emptymarkB{\ifx #1\@emptymarkA}%
\newcommand{\IFF}[1]{\mbox{$\equiv$}\ }
\newcommand{\IGNORE}[1]{{}}
\newcommand{\ILVALUE}[5]{\DENOTATION{#1}\mbox{$^{\mbox{#2,#3,#4,#5}}$}}

\newcommand{\IN}{\mbox{$\in$}\ }

\def\INA#1{\index{ZZZ-#1}}

\newcommand{\INFIXEXP}[3]{{[\TERM{#2} #1 \TERM{#3}]}}
\newcommand{\INMODEL}[1]{{\underline{#1}}}

\newcommand{\KEESIN}[1]{#1}
\newcommand{\KEESOUT}[1]{}
\newcommand{\LAMBDA}{\mbox{$\lambda$\/}}
\newcommand{\LAMBDAEXP}[2]{{\LAMBDA \TERM{#1}. #2}}
\newcommand{\LANGLE}{\mbox{$\langle$}}

\newcommand{\LEXITEM}[1]{{\em #1}}
\newcommand{\LOGEXP}[1]{{#1}}

\newcommand{\MDVALUE}[1]{\DENOTATION{#1}\mbox{$^{\mbox{M,d}}$}}
\newcommand{\MGDVALUE}[1]{\DENOTATION{#1}\mbox{$^{\mbox{M,g,d}}$}}

\newcommand{\NEQ}{\mbox{$\neq$}\ }
\newcommand{\NEWTERM}[1]{{\sc #1}}
\newcommand{\NLP}{\SIGLA{nlp}}
\newcommand{\NOT}{\mbox{$\neg$}}
\newcommand{\NP}{\SIGLA{np}}
\newcommand{\NOTE}[1]{}

\newcommand{\OR}{\mbox{$\vee$ }}
\newcommand{\PAIR}[2]{\LANGLE{\em #1},{\em #2\/}\RANGLE}
\newcommand{\PAR}[1]{\mbox{$\dot{#1}$}}
\newcommand{\PHI}{\mbox{$\Phi$}}
\newcommand{\PRED}[1]{{\bf #1}}
\newcommand{\PRIME}{\mbox{$'$}}
\newcommand{\PROBLEM}[1]{}
\newcommand{\PSBRACKET}[2]{{[\mbox{$_{\mbox{{\small #1}}}$} #2]}}
\newcommand{\PSI}{\mbox{$\Psi$}}
\newcommand{\QEXP}[4]{(#1 \TERM{#2}:\ #3 (#4))}
\newcommand{\QLF}{\SIGLA{qlf}}
\newcommand{\QUESTION}[1]{}
\newcommand{\RANGLE}{\mbox{$\rangle$}}
\newcommand{\RIGHTARROW}{\mbox{$\rightarrow$} }
\newcommand{\SAD}{\SIGLA{sad}}
\newcommand{\SECREF}[1]{\S\ref{#1}}
\newcommand{\SECTION}[2]{\section{#1}\label{#2}}
\newcommand{\SENTENCE}[1]{{\em #1}}
\newcommand{\SEQUENCE}[1]{{\LANGLE#1\/\RANGLE}}
\newcommand{\SHORTCITE}[1]{\shortcite{#1}}
\newcommand{\SHORTVALUE}[1]{\DENOTATION{#1}\mbox{$^{\mbox{M,d}}$}}
\newcommand{\SIGLA}[1]{{\small\uppercase{#1}}}
\newcommand{\SIGMA}{\mbox{$\sigma$}}

\newcommand{\SITUATIONS}{\mbox{${\cal S}$}}
\newcommand{\SOLUTION}[1]{}
\newcommand{\SREF}[1]{(\ref{#1})}
\newcommand{\SREFA}[2]{(\ref{#1}#2)}
\newcommand{\STILVALUE}[1]{\ILVALUE{#1}{M}{g}{w}{t}}
\newcommand{\STOREAWAY}{{\bf storeaway}}
\newcommand{\STORECOMBINE}[3]{{#1\mbox{$\otimes_{\mbox{#3}}$}#2}}

\newcommand{\SUBSECTION}[2]{\subsection{#1}\label{#2}}
\newcommand{\SUBSETEQ}{\mbox{$\subseteq$} }

\newcommand{\SUPSET}{\mbox{$\supset$ }}
\newcommand{\TAU}{\mbox{$\tau$}}
\newcommand{\TDAPPLY}[3]{{#1\{#2\}\mbox{$_{\mbox{#3}}$}}}
\newcommand{\TERM}[1]{{\em #1}}

\newcommand{\TRAINS}{\SIGLA{trains}}
\newcommand{\TRIN}[1]{#1}
\newcommand{\TROUT}[1]{}
\newcommand{\TYPE}[1]{\mbox{\it #1\/}}
\newcommand{\TYPEDPAR}[2]{\mbox{$\dot{#1}_{#2}$}}
\newcommand{\TYPES}{{\cal T}}
\newcommand{\UCOND}[2]{\PRED{#1}(\TERM{#2}\/)}
\newcommand{\UNION}{\mbox{$\cup$} }
\newcommand{\UNIVERSE}{\mbox{${\cal U}$}}
\newcommand{\VALUE}[2]{\DENOTATION{#1}\mbox{$^{\mbox{#2}}$}}
\newcommand{\VP}{\SIGLA{vp}}

\def\@AMPERSAND {&}
\newcommand{\@IFNOTBLANK}[2]{{\ifx #1\ \relax\else #2\fi}}
\newcommand{\DRTRULE}[7]{     
\begin{center}
\fbox{
\begin{tabular}{lc}
\multicolumn{2}{l}{#1} \\ \hline\hline\\[6ex]
\mbox{\begin{minipage}{40pt}
{\bf Triggering}\\
{\bf configuration}\\
\mbox{$\gamma$:}
\end{minipage}}                  & #2 \\[6ex]
\@IFNOTBLANK{#3}{{\bf Constraints:}} &
      \@IFNOTBLANK{#3}{#3}
      \ifx #4\ \relax\else \\[6ex]\fi
\@IFNOTBLANK{#4}{\mbox{\begin{minipage}{60pt}
                 {\bf Introducing}\\
                 {\bf into U$_{K}$:}
                 \end{minipage}}}
      \catcode `\& = 9
      \if  #4\ \relax \else \@AMPERSAND\fi
      \catcode `\& = 4
      \@IFNOTBLANK{#4}{#4}
      \ifx #5\ \relax\else \\[6ex]\fi
\@IFNOTBLANK{#5}{\mbox{\begin{minipage}{60pt}
                 {\bf Replace $\gamma$}\\
                 {\bf with:}
                 \end{minipage}}}
      \catcode `\& = 9
      \if  #5\ \relax \else \@AMPERSAND\fi
      \catcode `\& = 4
      \@IFNOTBLANK{#5}{#5}
      \ifx #6\ \relax\else \\[6ex]\fi
\@IFNOTBLANK{#6}{\mbox{\begin{minipage}{60pt}
                 {\bf Introduce}\\
                 {\bf into Con$_{K}$:}
                 \end{minipage}}}
      \catcode `\& = 9
      \if  #6\ \relax \else \@AMPERSAND\fi
      \catcode `\& = 4
      \@IFNOTBLANK{#6}{#6}
      \\[6ex]
#7
\\ 
\end{tabular}
}
\end{center}}

\makeatother

\begin{document}

\title{Semantic Ambiguity and Perceived Ambiguity\thanks{This paper will appear
in K. van Deemter and S. Peters (eds), {\em Semantic Ambiguity and
Underspecification}, CSLI.}}
\author{Massimo Poesio\\
University of Edinburgh\\
Centre for Cognitive Science\\
poesio@cogsci.ed.ac.uk}
\date{}

\maketitle

\begin{abstract}
I explore some of the issues that arise when trying to establish a connection
between the underspecification hypothesis pursued in the NLP literature and
work on ambiguity in semantics and in the psychological literature.  A theory
of underspecification is developed `from the first principles', i.e., starting
from a definition of what it means for a sentence to be semantically ambiguous
and from what we know about the way humans deal with ambiguity. An
underspecified language is specified as the translation language of a grammar
covering sentences that display three classes of semantic ambiguity: lexical
ambiguity, scopal ambiguity, and referential ambiguity. The expressions of this
language denote sets of senses. A formalization of defeasible reasoning with
underspecified representations is presented, based on Default Logic.  Some
issues to be confronted by such a formalization are discussed.
\end{abstract}

\section{The Combinatorial Explosion Puzzle}

The alternative syntactic readings of a sentence such as \SREF{synamb} probably
number in the hundreds, whereas sentences such as \SREF{hobbs} would have
hundreds of thousands scopally distinct readings if all permutations of
scope-taking sentence constituents were considered admissible readings. Yet,
human beings appear able to deal with these sentences effortlessly.

\begin{EXAMPLE}
\ENEW{synamb} We should move the engine at Avon, engine E1, to Dansville to
	pick up the boxcar there, then move it from Dansville to Corning,
        load some oranges, and then move it on to Bath.
\ENEW{hobbs} A politician can fool most voters on most issues most
	of the time, but no politician can fool all voters on every
        single issue all of the time.
\end{EXAMPLE}
This \NEWTERM{Combinatorial Explosion Puzzle} is one of the most fundamental
questions to be addressed by a theory of language processing, and a substantial
problem for developers of Natural Language Processing (\NLP) systems.  {\NLP}
systems which have to perform non-linguistic actions like booking a flight in
response to an user's utterance must arrive at the preferred interpretation of
their input in the context of the conversation, if one exists; otherwise, they
must realize that their input is ambiguous and request a clarification.
Examples such as \SREF{synamb} and \SREF{hobbs} indicate that such systems
cannot adopt the sentence processing strategy of generating all the readings of
an ambiguous sentence and choosing one of them, because there are too many such
readings.\TRIN{\footnote{With current technology, the real problem is not so
much the size of the search space, but how to choose among these
interpretations, most of which are plausible.}} In order to develop such
systems, a theory of ambiguity processing is needed that is consistent both
with linguistic facts and with what is known about the way humans disambiguate.

\begin{sloppypar}
Work on \NEWTERM{underspecified representations} such as
\cite{alshawi:CLE-book,poesio:STASS2,reyle:93}\TRIN{\footnote{An early attempt
at a model of discourse interpretation of this kind was made by Hobbs, e.g.,
\cite{hobbs:83}.}} differs from other work on discourse interpretation
{}~\cite{charniak&goldman:88,hobbs-et-al:abduction-TR,%
pereira&pollack:incremental,dalrymple-et-al:91,hwang&schubert:STASS3,%
kamp&reyle:93} because it is explicitly motivated by the Combinatorial
Explosion Puzzle, and aims at a unified account of all interpretation
processes, including those that occur before the scope of all operators has
been determined. The work on underspecified representations holds the promise
of yielding a better account of the way interpretive processes such as scope
disambiguation and reference resolution affect each other.
\end{sloppypar}

The existing theories of underspecification, however, have been motivated
almost exclusively by computational considerations. For example, the semantics
assigned to underspecified representations is designed so as to support those
inferences that are deemed useful for an economical approach to disambiguation,
rather than being motivated by an analysis of the phenomenon of ambiguity.  In
this paper I explore some of the issues that arise when trying to establish a
connection between work on underspecification and, on the one side, work on
ambiguity in semantics; on the other side, work on ambiguity in the
psychological literature. A theory of underspecification is developed `from the
first principles', i.e., starting from a definition of what it means for a
sentence to be semantically ambiguous and from what we know about the way
humans deal with ambiguity. The goal is to arrive at a linguistically and
cognitively plausible theory of ambiguity and underspecification that, in
addition to computational gains, may provide a better understanding of how
humans process language.\TRIN{\footnote{A recent example of work also
attempting to exploit the properties of underspecified representations to
address linguistic questions is \cite{crouch:EACL95}.}}

Many of the issues discussed in this paper arose from work on the {\TRAINS}
project at the University of Rochester, in which the issues of language
comprehension, planning, and reasoning encountered in task-oriented natural
language conversations are studied \cite{allen-et-al:journal}. The theory of
ambiguity proposed in this paper is the basis for the implemented surface
discourse interpretation system {\SAD}-93, used in the {\TRAINS}-93 demo
system. {\SAD}-93 is described in \cite{poesio:thesis}.

\SECTION{Ambiguity in Natural Language}{ambiguity_section}

\subsection{Ambiguity and Grammar}

\TRIN{
\subsubsection{Characterizing Ambiguity}

The dictionary definitions of the terms \NEWTERM{ambiguity} and
\NEWTERM{ambiguous} try to capture the intuition that an expression is
ambiguous if `it has multiple meanings'. An example are the following entries,
from Webster's:

\begin{itemize}
  \item
   am.big.u.ous $|$am-'big-y*-w*s$|$ aj [L ambiguus, fr. ambigere to
   wander about, fr. ambi- + agere to drive -- more at agent]

   \ldots

   2: capable of being understood in two or more possible senses
   : equivocal - am.big.u.ous.ly av
  \item
   am.bi.gu.ity $|$am-b*-'gy{u:}-*t-{e-}$|$ n : the quality or
   state of being ambiguous in meaning; also : an ambiguous word
   or expression
\end{itemize}
A more precise
characterization of the notion of ambiguity is required to differentiate
ambiguity from \NEWTERM{vagueness} or \NEWTERM{indeterminacy}, for example, or
to clarify notions such as \NEWTERM{homonimy} and \NEWTERM{polysemy} (see
below).\footnote{A sentence is \NEWTERM{indeterminate}, or
\NEWTERM{unspecified}, if it is definitely true or false, but it could be made
more specific.  Zwicky and Sadock \shortcite{zwicky&sadock:75} bring the
example of the sentence \SENTENCE{My sister is the Ruritanian secretary of
state}, which is indeterminate as to whether ``\ldots my sister is older or
younger than I am, whether she acceeded to her post recently or some time ago,
whether the post is hers by birth or by merit,'' and so forth. The point is
that these additional facts do not affect the truth value of the sentence. It
hardly needs to be pointed out that just about every sentence is indeterminate
/ unspecific in some respects. I will use the term indeterminate for these
sentences, and reserve the term \NEWTERM{underspecified} for sentences which
may have different truth values depending on the way the facts are `filled in'
(see below).}

An early attempt at making the notion of ambiguity more precise was presented
in \cite{lakoff:vagueness-ambiguity}. Lakoff proposed linguistic tests that
could be used to tell whether a sentence was ambiguous or
not.\TRIN{\footnote{An example of these tests are the \NEWTERM{identity tests},
one of which is the \NEWTERM{conjunction} test. The (presumed) ambiguity of
sentences such as \SENTENCE{They say her duck} derives from the fact that the
phrase \SENTENCE{her duck} can either be a NP or a bare infinitival
complement. The sentence \SENTENCE{They saw her swallow} should have a similar
ambiguity. If these two sentences were really ambiguous as claimed, a sentence
such as \SENTENCE{They saw her swallow and her duck} should only have two
readings instead of four since conjunction requires its two arguments to be of
the same type, and therefore the 'crossed' readings should not be
available. This is indeed the case. On the other hand, an indeterminate
sentence such as \SENTENCE{My sister is the Ruritanian secretary of state}
maintains all of its indeterminateness once conjoined, as in \SENTENCE{My
sister is the Ruritanian secretary of state and a prominent composer}.}}
(Lakoff's tests were meant to provide a way for distinguishing ambiguous
sentences from indeterminate ones.)  Zwicky and Sadock
\shortcite{zwicky&sadock:75} showed, however, that such tests do not result in
an unambiguous classification of sentences, and that a formal characterization
of the concepts of ambiguity and indeterminacy was required even to understand
what these `ambiguity tests' really test.\TRIN{\footnote{For example, Zwicky
and Sadock observed that such tests can only identify \NEWTERM{polar}
ambiguities (such as the ambiguity of \NEWTERM{game} between two entirely
independent readings), but not \NEWTERM{privative} ambiguities, like those
displayed by a term like \NEWTERM{dog} which can be used both to indicate
generic individual of the Canis species and a male element of it.}}

}

\subsubsection{Meaning, Sense, and Ambiguity}

One problem to be tackled in attempting to make precise the definition of
ambiguity is to say what `meanings' and `senses' are.  In modern semantic
theory, the meaning assigned to an expression by a grammar is a function from
contexts (or \NEWTERM{discourse situations}) to \NEWTERM{senses}.  Roughly
speaking, the discourse situation provides a value for all context-dependent
aspects of the sentence; the sense of a sentence (what we get once we resolve
its context-dependent aspects) tells us under which circumstances in the world
the sentence is true or false.\footnote{For a discussion of these assumptions,
see \cite{kaplan:77}, \cite{BarPer:saa}, or chapter 2 of \cite{pinkal:lal}.}

Not all notions of `sense' employed in the literature can serve as the basis
for a definition of ambiguity. For example, of the various notions of
\NEWTERM{proposition} (the sense of sentences), the simplest is the one
according to which propositions are truth values. But if we were to use this
notion of sense, the sentence \SENTENCE{Kermit croaked}, ambiguous between a
reading in which Kermit utters a frog-like sound and a reading in which he
dies, would be classified as unambiguous with respect to all models in which
Kermit has both the property of dying and the property of producing a frog-like
sound, or he (it) has neither property. In other words, in providing a
definition of ambiguity we find the same need for a fine-grained notion of
sense that has been observed in connection with the semantics of attitude
reports.\footnote{In the case of attitude reports, the problem is to make sure
that if \SENTENCE{John is tall} and \SENTENCE{John is stupid} are both true in
a model, \SENTENCE{Bill believes that John is tall} does not entail
\SENTENCE{Bill believes that John is stupid} in that model, assuming that
propositions are the semantic correlate of sentential complements.} A
model-theoretic definition of ambiguity requires a finer-grained notion of
proposition than simply truth values. In most recent semantic theories, senses
are intensional objects; the simplest way of achieving intensionality is to use
functions from possible worlds or situations to referents as one finds in
Montague Grammar, where, for example, propositions are functions from possible
worlds to truth values. This simple form of intensionality will be sufficient
for the purposes of the present paper.\footnote{More complex notions of
propositions have been introduced in the literature on propositional attitudes,
such as those used in Situation Semantics \CITE{BarCoo:ekn} or Property
Theory \CITE{Turner:ppast}.}

\subsubsection{A Semantic Theory of Ambiguity}

The notions of `meaning' and `sense' just discussed are the starting point for
the semantic account of the notion of ambiguity and its relation with vagueness
developed by Pinkal (\shortcite{pinkal:lul}, translated as \cite{pinkal:lal}).
Pinkal introduces the notion of \NEWTERM{indefiniteness} to subsume both
ambiguity and vagueness. He defines indefiniteness as follows:

\begin{DEFINITION}
  A sentence is \NEWTERM{semantically indefinite}
  if and only if in certain situations,
  despite sufficient knowledge of the relevant facts, neither ``true'' nor
  ``false'' can be clearly assigned as its truth value.
\end{DEFINITION}
Pinkal formalizes the notion of indefiniteness in terms of
\NEWTERM{precisification}.\footnote{A treatment of ambiguity and vagueness in
terms of precisifications was proposed early on in \cite{fine:75}.}  According
to Pinkal, a linguistic expression is semantically indefinite if it has the
potential for being made {\em precise} in distinct ways. For example, the
sentence \SENTENCE{The Santa Maria is a fast ship} containing the degree
adjective \LEXITEM{fast} can be `made precise' (and assigned a definite truth
value) either with respect to a context in which `fast' is interpreted as `fast
for a modern ship', in which case the sentence is false; or with respect to a
context in which `fast' is interpreted as `fast for a ship of her age', in
which case the sentence can be true or false, depending on the class of
comparison.  Let $p$ and $q$ be two propositions. Proposition $p$ is
\NEWTERM{more precise than} $q$ iff (i) $p$ is true (false) under all states of
the world under which $q$ is true (false), and (ii) $p$ is true or false under
certain circumstances under which $q$ is indefinite. The idea of
\NEWTERM{precisification} is defined as follows:

\begin{DEFINITION}
  Expression {\ALPHA} in context \TERM{c} \NEWTERM{can be precisified}
  to \TERM{s} if and only if (i) \TERM{s} is a sense that {\ALPHA} can
  assume according to its meaning; and (ii) \TERM{s} is more precise than
  the sense of {\ALPHA} in \TERM{c}.
\end{DEFINITION}
The connection between indefiniteness and precisification is provided by the
following \NEWTERM{Precisification Principle}:

\begin{description}
\item [Precisification Principle]:
  A sentence is of indefinite truth value in a context if and only if
  it can be precisified alternatively to ``true'' or to ``false''.
\end{description}
which Pinkal also reformulates as follows:
\begin{description}
  \item [Extended Precisification Principle] An expression is semantically
indefinite in a context iff it can assume different senses in that context.
\end{description}
Pinkal does not equate ambiguity with vagueness.  His theory includes,
in addition to the notion of precisification, additional criteria to
differentiate different forms of ambiguity, as well as differentiating `pure'
ambiguity from `pure' vagueness. The intuition he is trying to capture is that
``\ldots whether an expression is ambiguous or only vague is a question that
cannot be cleared once and for all. Indefiniteness is perceived as ambiguity
when alternative precisifications are predominant, as vagueness when an
unstructured continuum presents itself:''

\begin{description}
\item [Ambiguity (Pinkal)]:
  If the precisification spectrum of an expression is perceived as discrete,
  we may call it ambiguous; if it is perceived as continuous, we may call
  it vague.
\end{description}
Pinkal identifies two fundamental
types of ambiguity, according to whether an expression has, or does not have, a
`wider' sense that could be taken as most `basic'. For example, \NEWTERM{ball}
does not have a wide sense of `round object and dancing party', whereas
\NEWTERM{American} may either mean `person from the US' or `person from the
American continent'. He classifies expressions like \NEWTERM{American}
which have a wider sense as having a \NEWTERM{multiplicity of use}, whereas
expressions such as \NEWTERM{ball} or \NEWTERM{green} which do require
precisification are called \NEWTERM{narrowly ambiguous}. The cases of ambiguity
in the narrow sense are further distinguished in two classes, depending on
whether they are subject to the \NEWTERM{Precisification Imperative}. Although
the two interpretations of \NEWTERM{green} are distinct, it is possible of an
object to be both green in the `ripe' sense and green in the `color' sense: for
example, a green apricot. An object cannot, however, be a `band' both in the
musical group sense and in the piece of tape sense. Pinkal proposes that
polysemous expressions behave like \NEWTERM{green}, and calls all of these
expressions \NEWTERM{P-type ambiguous}; expressions like \NEWTERM{band},
however, are true homonyms, and therefore he calls them \NEWTERM{H-type
ambiguous}. These latter are defined as follows:

\begin{description}
  \item [Precisification Imperative]: An expression is H-type ambiguous
  iff its base level is inadmissible, i.e., if it requires precisification.
\end{description}
For my purposes,  it's
not particularly important whether the difference between homonimy and polysemy
is completely captured by the Precisification Imperative; what is important is
the claim that H-type ambiguous expressions need precisification, and
furthermore, that the Precisification Imperative ``is a second order phenomenon
\ldots that lies beyond the scope of a strictly truth-conditional approach.''
(\cite{pinkal:lal}, p. 86--87). I will provide below independent reasons for
including a formalization of reasoning in context in a treatment of ambiguity,
and I will argue that such formalization provides the necessary tools to
express the Precisification Imperative.

To summarize, a sentence is H-type ambiguous iff the grammar assigns to it
distinct precisifications (senses) in a given discourse situation, and if the
`base level' of the expression requires precisification.  Thus, the sentence
\SENTENCE{Kermit croaked} is considered ambiguous since in the `empty context'
that provides all the senses of the expression according to the grammar G for
English, that sentence has two senses: the proposition that attributes to
Kermit the property of producing the sound that frogs produce, and the
proposition that attributes to Kermit the property of dying.  (I am assuming
here that terms like \LEXITEM{Kermit} refer unambiguously.) On the other hand,
the sentence \SENTENCE{Kermit kissed Miss Piggy} would be considered
unambiguous with respect to the same context.

Although Pinkal is only concerned with lexical ambiguity, the precisification
approach can also be used to classify as ambiguous sentences which have more
than one structural analysis (like the sentence \SENTENCE{They saw her duck})
or are scopally ambiguous (cfr. the sentence \SENTENCE{Everybody didn't leave})
whenever the grammar assigns to them more than one sense. I will discuss below
how Pinkal's system can be extended to scopal and referential
ambiguity.\TRIN{\footnote{Another way of providing a precise definition of
ambiguity has been explored in the literature, that we might call {\em
structural} or {\em syntactic}. An example of structural definition is the
following, from Gillon (\cite{gillon:synthese}, p. 400):

\begin{description}
  \item [Ambiguity (Gillon)]: {\em An expression is ambiguous iff the
expression can accomodate more than one structural analysis.}
\end{description}
This definition relies on the assumption made in transformational theories of
grammar such as Government and Binding theory
\CITE{chomsky:LGB,haegeman:book}, that each interpretation of a sentence is a
quadruple $\langle$PF,DS,SS,LF$\rangle$, each of whose elements is a structured
object: PF characterizes the phonetic interpretation, DS the predicate/argument
composition of the sentence, SS its surface syntactic analysis, and LF its
logical analysis.  A sentence (string) is ambiguous iff it can be characterized
by distinct quadruples, and this may happen not only for phonetical or
syntactic reasons, but also for semantical reasons, since one of the components
of a structural characterization of a sentence, the `LF,' encodes the semantic
interpretation of the sentence.

The problem with this definition is that a purely syntactic analysis of meaning
introduces spurious distinctions: for example, unless something is said about
invariance under renaming of variables, one would predict from a structural
definition that even a sentence with a single quantifier such as
\SENTENCE{Every man left} is infinitely ambiguous, because all structures of
the form \PSBRACKET{S}{\PSBRACKET{NP}{every x man}\PSBRACKET{x}{left}}, for any
choice of the variable, are appropriate (and distinct) LF constituents of a
sentence's interpretation. Another case of spurious ambiguity is discussed
below.  A semantic characterization of ambiguity avoids this problem.}}

\subsubsection{The Disjunction Fallacy}

It is important to realize that saying that a sentence is ambiguous in a
context if it has distinct precisifications is not the same as saying that an
ambiguous sentence is equivalent to the disjunction of its distinct
precisifications. Intuitively, in uttering S, whose two precisifications are
the propositions P and Q, a speaker may have meant P or she may have meant Q,
but the following does not hold:

\begin{center}
[[A \PRED{means} that P] \OR [A \PRED{means} that Q]] \EQUIV
[A \PRED{means} that [P \OR Q]]
\end{center}
To treat an ambiguous sentence in such a way would be tantamount to propose
that an ambiguous sentence has a single sense in any given discourse situation,
namely, the proposition that is true at a situation if either of the distinct
interpretations of the sentence is true at that situation; but according to the
definition above, an ambiguous sentence is one which has more than one sense at
a discourse situation.  For example, according to the definition of ambiguity
discussed above, the listener of an utterance of \SENTENCE{They saw her duck}
could either interpret the speaker as saying that the contextually determined
set of individuals denoted by the pronoun \LEXITEM{they} saw a contextually
specified female person lowering herself, or as saying that that set of
individuals saw the pet waterfowl of that female person. According to the
disjunction theory, instead, the listener would attribute to the speaker of
that sentence a single meaning, albeit a disjunctive one; namely, that it was
either the case that \LEXITEM{they} saw a contextually specified female person
lowering herself, or it was the case that \LEXITEM{they} saw the pet waterfowl
of that female person.\footnote{A cute example of the problems with the theory
is presented by  \cite{stallard:87}. If ambiguous sentences were to
denote the disjunction of their readings, then the answer to the question
\SENTENCE{Does the butcher have kidneys?} should always be 'yes'.}

I will refer to the idea that a semantically ambiguous sentence denotes the
disjunction of its alternative interpretations as the \NEWTERM{disjunction
fallacy}. The disjunction fallacy can be found in the literature in two forms.
Its `purest' form is the hypothesis that the interpretation process literally
involves generating all of the senses of an expression and putting them
together in a disjunction. In this form, the disjunction 'theory' is not simply
counterintuitive; it doesn't explain the combinatorial explosion puzzle at
all. As far as I know, this `explicit' form of the theory has only been
discussed jokingly.\FOOTNOTE{CHECK HOBBS REFERENCE FROM STALLARD.} One can find
in the literature, however, an `implicit' form of the disjunction theory, in
theories of underspecification that assign to underspecified expressions a
semantics that makes them equivalent to the disjunction of their readings. One
such proposal is \cite{poesio:STASS2}; the semantics of UDRSs is also
disjunctive \CITE{reyle:93}.

\TRIN{

\SUBSECTION{The Role of Syntactic and Semantic
Constraints}{syn_sem_constr_section}

Although the number of logical form permutations that one can obtain for a
particular sentence by, e.g., considering all the permutations of its operators
may be rather large, constraints of a syntactic and/or semantic nature
drastically reduce this number.

In the case of scopal ambiguity, for example, permutations may not correspond
to actual readings for at least three reasons. First of all, some of these
permutations result in logical expressions that are either ill-formed or
contradictory,as noted by, e.g., Hobbs an Shieber \shortcite{hobbs&shieber:87}.
For example, \SREFA{samb-ex:impossible}{a}, in the interpretation in which the
pronoun \LEXITEM{he} is anaphoric on the {\NP} \LEXITEM{every man,} does not
have a reading in which the {\NP} \LEXITEM{the woman he$_i$ married} outscopes
the {\NP} \LEXITEM{every man$_i$}. There is no well-formed logical expression
that may represent this reading.  Hobbs and Shieber point out that this
constraint also prevents a quantifier to scope between a noun and its
complement: for example, \LEXITEM{a meeting} may not scope inside
\LEXITEM{most} but outside \LEXITEM{each} in
\SREFA{samb-ex:impossible}{b}.\TRIN{\footnote{Pereira
\shortcite{pereira:CL90} argues that this constraint is best formulated as a
condition on semantic derivations rather than as a condition on the syntax of
logical expressions.}}

\begin{EXAMPLE}
\ENUMA{samb-ex:impossible}
\EITEM Every man$_i$ loves  the woman he$_i$ married.
\EITEM Most people on each committee attended
		a meeting.
\ENDENUMA
\end{EXAMPLE}
Another reason why the number of actual readings of a sentence is much
smaller than the number of permutations of its operators is that two distinct
permutations may correspond to semantically equivalent readings. For example,
\SREF{samb-ex:4} has only one reading, even though (at least) two equivalent
logical expressions can be obtained as the translation of the
sentence.\footnote{This is one of the reasons for preferring a semantic account
of ambiguity to a syntactic account which makes ambiguity depend on the
existence of two distinct logical forms.}


\begin{EXAMPLE}
\ENEW{samb-ex:4} A student saw a dog.
\end{EXAMPLE}
Finally, the readings corresponding to certain permutations may be
unavailable because of syntactic constraints.  Much work on uncovering readings
that are absent due to constraints on syntactic trasformations and/or
conditions on syntactic levels of representation has been done in the
generative tradition \cite{may:85}.

Some of the constraints proposed in this literature have been proved to yield
quite robust predictions.  Perhaps the best known example of syntactic
constraint is the observation that a quantifier cannot take scope outside the
clause in which it appears.  The observation that clauses serve as `scope
islands' goes back at least to Rodman \shortcite{rodman:76}, but was discussed
most extensively by May ~\SHORTCITE{may:85}; the constraint was called
\NEWTERM{Scope Constraint} by Heim \SHORTCITE{heim:82}. The Scope Constraint
is exemplified by the contrast in \SREF{sc-1}: whereas \SREFA{sc-1}{a} has a
reading in which \LEXITEM{every department} is allowed to take wide scope over
\LEXITEM{a student,} this reading is not available for \SREFA{sc-1}{b}, even
though arguably \LEXITEM{from every department} and \LEXITEM{who was from every
department} have the same denotation.
\begin{EXAMPLE}
\ENUMA{sc-1}
\EITEM A student from every department was at the party.
\EITEM A student who was from every department was at the party.
\ENDENUMA
\end{EXAMPLE}
Although syntactic and semantic constraints do not rule out all possible
readings---for example, a sentence like \SENTENCE{They saw her duck} still has
more than one interpretation under all of these theories---a theory of
disambiguation must be such that these constraints can play a role.

}

\subsection{Perceived  Ambiguity}

As noted by Hirst \SHORTCITE{hirst:ambiguity-book}, the discussions of
ambiguity processing in the {\NLP} literature tend to ignore the fact that
humans are aware that sentences can be ambiguous, and that they can exploit the
ambiguity of sentences for rhetorical effect.  Raskin, for example, claims
\SHORTCITE{raskin:humor} that humor crucially relies on ambiguity. He
discusses examples such as the following (p.  25-26):

\begin{EXAMPLE}
\ENEW{raskin:2} The first thing that strikes a stranger in New York is a
	big car.
\end{EXAMPLE}
The joke relies on two assumptions about human processing: first, that
the clause \LEXITEM{the first thing that strikes a stranger in New York} gets
interpreted before the end of the sentence, with \LEXITEM{strikes} receiving
the `surprise' interpretation; and second, that the reader, upon reading
\LEXITEM{is a big car}, will go back, produce a second interpretation, and
entertain both interpretations simultaneously. The joke could not be understood
unless the hearer were able to entertain the two interpretations of the
sentence simultaneously. These jokes can exploit other forms of ambiguity,
e.g., scopal ambiguity, as in \SENTENCE{Statistics show that every 11 seconds a
man is mugged here in New York City. We are here today to interview him.}

The reader's ability to entertain more than one interpretation simultaneously
is exploited in poetry, as well \CITE{su:1994}.  The linguistic articles
discussing ambiguity are another literary form that exploits this
possibility. Examples such as \SENTENCE{They saw her duck} are a clear case of
deliberate ambiguity; the whole point of these examples is to show that a
sentence can have more than one interpretation. The writer relies on the reader
being able to entertain more than one interpretation at once.

\TRIN{
The opposite is true, as well: when clarity is a goal, writers and speakers
tend to construct their sentences in such a way as to avoid ambiguity. Thus,
most sentences one runs across in scientific texts or in transcripts of
task-oriented conversations have a clearly preferred interpretation.  This
interpretation is sometimes suggested by the context, sometimes by means of
\NEWTERM{disambiguation markers}---expressions such as \LEXITEM{each},
\LEXITEM{a different}, or \LEXITEM{the same} that suggest which interpretation
is preferred. Thus, a writer will use sentences such as \SENTENCE{Every kid
climbed the same tree}, rather than \SENTENCE{Every kid climbed a tree}, when
we/she wants to make sure that the reader arrives at the interpretation in
which there is a single tree.\footnote{Such sentences are used, for example, to
get a 'baseline' interpretation in psychological work on ambiguity.}
}

I will call the situation in which a listener arrives at more than one
interpretation for an utterance \NEWTERM{perceived ambiguity}.  A situation in
which B perceives an utterance as ambiguous may result in B's appreciating the
joke, the poetic phrase, or the point of the linguistic example; if the
ambiguity is not perceived as intended, B may say saying something like
\SENTENCE{This is not very clear,} or perhaps \SENTENCE{This sentence is
ambiguous}. This situation can be informally characterized as follows:

\begin{DEFINITION}
An utterance U by conversational participant A addressing
conversational participant B in a discourse situation D is \NEWTERM{perceived
as ambiguous in D} by B if B's processing of U in D results in B obtaining
distinct interpretations for U.
\end{DEFINITION}
\begin{sloppypar}
The phenomenon of deliberate ambiguity  suggests that the solution to the
Combinatorial Explosion Puzzle cannot be that humans either generate only one
interpretation at a time by using some clever heuristics, or do not generate
any interpretation at all. Humans entertain more than one interpretation at a
time, and they may not be able to choose one among them. This conclusion is
also supported by psychological results. There is evidence, for example, that
during both lexical processing and syntactic processing several hypotheses are
generated in parallel, and only later filtered on the basis of contextual
information
\CITE{swinney:79,crain&steedman:85,kurtzman:thesis,schubert:attachment,%
gibson:thesis}.\footnote{The results about lexical disambiguation are fairly
well established, but there is some controversy about syntactic processing. A
constrasting view on syntactic disambiguation is discussed in
\cite{frazier&fodor:78}.} Kurtzman and MacDonald
\SHORTCITE{kurtzman&macdonald:93} suggest a similar model for scope
disambiguation. As far as reference interpretation is concerned, there is some
evidence that all pragmatically available referents become active before a
referent is identified (see, e.g., \cite{spivey-et-al:AAAI94}).
\end{sloppypar}

These facts are consistent with the view of discourse interpretation taken in
Artificial Intelligence, in which processes such as reference resolution or
lexical disambiguation are modeled in terms of defeasible inference, which may
result in alternative hypotheses. Examples include the theories of the effects
of semantic priming on lexical disambiguation, as formalized, e.g., in Hirst's
{\small ABSITY} system \CITE{hirst:ambiguity-book} or, more recently, in
statistically based terms; the theories about the effects of local focusing on
the choice of pronoun antecedents such as \cite{GJW:83}; the work on temporal
interpretation by Asher, Lascarides, and Oberlander (see, e.g.,
\cite{lascarides-et-al:ACL92}); and work on scopal disambiguation such as
\cite{kurtzman&macdonald:93,poesio:thesis}.

\subsection{Semantic Ambiguity versus Perceived Ambiguity}

A preliminary and, I hope, uncontroversial conclusion I intend to draw from the
discussion on deliberate ambiguity and ambiguity processing is that a theory of
ambiguity that aims at explaining the Combinatorial Explosion Puzzle needs to
be concerned both with the interpretation that the grammar assigns to a
sentence---i.e., what it means for a sentence to be semantically
ambiguous---and with the process by which interpretations are generated, i.e.,
with what it means for an utterance to be perceived as ambiguous.  On the one
hand, the theory must explain why the disambiguation process will not generate
all semantically available interpretations; on the other hand, it must predict
that more than one interpretation will be generated. This conclusion is the
central idea of this paper, indeed, what gives the paper its title. The
inclusion of a theory of disambiguation will also remedy one of the omissions
in Pinkal's theory, namely, how to formalize the Precisification Imperative.

The discussion of perceived ambiguity supports a stronger claim, namely, that
semantic ambiguity and perceived ambiguity are distinct notions, in the sense
that whereas a model of semantic ambiguity has to express the truth-conditional
properties of an expression, the reasoning processes involved in
disambiguation, and that may lead to a perceived ambiguity, consist of
defeasible inferences that are not supported by the semantics of ambiguous
expressions.

The distinction I intend to draw, then, is as follows. Semantic ambiguity is
part of the specification of the grammar of a language; most, if not all,
sentences are semantically ambiguous, but their ambiguity need not be noticed
by listeners, and in fact it is typically discovered only by linguistic
research. Perceived ambiguity, on the other hand, is a result of the
interpretation process, that is defeasible in nature, and may therefore result
in more than one interpretation in cases of miscommunication or when the
speaker constructs the context appropriately to serve a rhetorical purpose, as
in the puns presented above.

Some readers may wonder why the developer of a {\NLP} system should be
concerned with perceived ambiguity, i.e., with generating all of the
contextually available interpretations of a sentence. The answer is that
certain applications need this information. Consider the following example,
again from the {\TRAINS} domain. Say that the user utters \SENTENCE{move the
engine to Avon}, and say that two different engines have been discussed during
the elaboration of the current part of the plan. Clearly, we do not want the
system to just come out with a plausible guess about which engine was meant:
instead, we want it to recognize the ambiguity and ask for clarification. In
general, all systems that engage in conversations with their users need to be
able to recognize an ambiguity, to ask for clarifications when necessary rather
than guess one possible interpretation, and to make their own output
unambiguous. (Of course, the theory of contextual disambiguation must be such
that no spurious ambiguities are obtained.)

\SECTION{The Underspecification Hypothesis}{undersp_section}

All theories of semantic interpretation based on Montague's general program as
exposed in {\em Universal Grammar} \CITE{montague:UG} assume that the grammar
of a language $\cal L$ specifies two homomorphisms: one between syntactic trees
and a \NEWTERM{disambiguated language} ${\cal DL}$, and a second one between
the disambiguated language and objects of the model M (the senses). These two
homomorphisms can be composed, thus making the intermediate level of the
disambiguated language dispensable. The grammar assigns to an ambiguous
expression of ${\cal L}$ distinct expressions of ${\cal DL}$, each of which has
a unique interpretation.

A direct implementation of this strategy in an {\NLP} system would require
generating all senses of an ambiguous sentence-string, which would be clearly
problematic.  Many {\NLP} systems, instead, make use of heuristic methods that
generate only one interpretation and ignore the alternatives. These heuristics
work fairly well fairly often; such systems, however, won't be able to perceive
an ambiguity even when it would be helpful to do so. Other systems therefore
split the semantic problem of computing all the interpretations of a sentence
from the processing problem of generating these interpretations in context, by
making use of an intermediate, \NEWTERM{underspecified} level of
representation.  One of the earliest examples of underspecified representations
is the `Logical Form' of Schubert and Pelletier
\shortcite{schubert&pelletier:82}. The representation for \SREF{kmcd:3}
proposed by Schubert and Pelletier, shown in \SREF{kmcd:3:ULF}, is a typical
example of these underspecified representations: quantifiers are left in place
and the referent for the definite description \LEXITEM{the tree} is not
specified.
\begin{EXAMPLE}
\ENEW{kmcd:3}     Every kid climbed the tree.
\ENEW{kmcd:3:ULF} [$<$every kid$>$ climbed $<$the tree $>$]
\end{EXAMPLE}
In more recent years,
underspecified representations similar to Schubert and Pelletier's have been
used by \cite{hobbs&shieber:87}, Fenstad {\ETAL} \shortcite{fenstad-et-al:87},
in Allen's textbook \SHORTCITE{allen:87} and, most recently, in the Core
Language Engine \CITE{alshawi:CLE-book}; the `uninterpreted conditions'
produced during the intermediate steps of the {\DRT} construction algorithm in
\cite{kamp&reyle:93} can be considered underspecified representations as well.

Underspecified representations were originally conceived as a way to solve a
problem in system implementation, namely, separating `context-independent' from
`context dependent' aspects of the interpretation, thus making either part
reusable for different applications.\KEESOUT{\footnote{In addition to the
references above see \cite{woods:scope}.}} Since the motivation was strictly
computational, the underspecified representations used in most {\NLP} systems
are little more than data structures, in the sense that they do not have a
interpretation other than the one provided by the procedures that interpret
them. These representations `encode' the ambiguity of a sentence in the sense
that that sentence has the reading r iff that reading can be generated by
repeatedly applying `construction rules' to the underspecified representation.

In recent years, there has been growing interest for the hypothesis that the
ability to encode multiple interpretations in an underspecified language may be
(part of) the explanation of the Combinatorial Explosion Puzzle. The idea is
that humans, as well, make use of an underspecified language that can encode
distinct meanings implicitly, and therefore do not need to generate all of
these meanings.  A semantically ambiguous sentence, therefore, need not cause
problems for a human to process, because it is not necessarily {\em perceived}
as ambiguous in the sense discussed in the previous section. I will call this
assumption the \NEWTERM{Underspecification Hypothesis}:

\begin{description}
  \item [Underspecification Hypothesis]:
Human beings represent semantic ambiguity implicitly by means of
\NEWTERM{underspecified representations} that leave some aspects of
interpretation unresolved.
\end{description}
My goal in the rest of the paper is to  spell out  the
Underspecification Hypothesis both as a theory of grammar and as a theory of
discourse interpretation. I assume, that is, that the hypothesis is correct,
and try to answer questions such as: what kind of language are underspecified
representations?  what is their semantics? and, what kind of inferences are
done with them?

The novel aspect of this work is that the answers I give are based on the
discussion of semantic ambiguity and perceived ambiguity in the previous
section.  I hypothesize that underspecified representations are used by humans
as the translation of expressions that are indefinite in the sense of Pinkal,
and assign them a semantics that reflects this hypothesis. I assume that the
disambiguation process is consists of defeasible inferences, and examine the
characteristics of defeasible reasoning with underspecified
representations. Although the same position towards disambiguation and
defeasible has been adopted in the Core Language Engine, most of the issues I
discuss have not been mentioned so far in the discussion on underspecified
representations.

In the literature on underspecification, one often finds the argument that
providing a semantics to underspecified representations is necessary because
disambiguation requires inference, and therefore a `logic of
underspecification' is needed
\CITE{poesio:STASS2,reyle:93,van-deemter:thesis}.  However, it is not at all
clear whether the process of disambiguation involves much semantically
justified reasoning; disambiguation seems to consist mostly of defeasible
inferences. It is fair to say that the debate on this issue is very open at the
moment, as certified by a number of recent panels on the subject.  But whatever
the final conclusion on this topic will be, it is clear that under the
perspective that the grammar of a language $\cal L$ is a mapping from elements
of $\cal L$ to underspecified representations, the semantics of these
underspecified representations becomes a central aspect of the specification of
the grammar. Furthermore, it also becomes clear that the semantics of
underspecified representations must be based on an analysis of semantic
ambiguity, otherwise we wouldn't even know whether the form of underspecified
representation we develop does the job it is supposed to do.

\SECTION{An Underspecified Theory of Ambiguity, Part I:  Lexical
	 Ambiguity}{lexamb_section}

The simplest way to illustrate my implementation of the Underspecification
Hypothesis is to start with lexical ambiguity. I present in this section a
theory of grammar which makes use of an underspecified language to encode the
`ambiguity potential' of lexically ambiguous expressions, as well as a simple
formalization of lexical disambiguation as defeasible inference over
underspecified representations. In the next section I will show how to extend
the approach presented here to deal with expressions that exhibit other forms
of semantic ambiguity.

I want to emphasize that I start with lexical disambiguation for expository
purposes only. Lexical ambiguity is the one case of ambiguity for which a
`generate and test' strategy may well be compatible with the psychological
results, therefore the one for which the need for underspecified
representations is less clear. Furthermore, I will only discuss cases of
lexical ambiguity in the narrow sense, which is perhaps the least interesting
case of lexical indefiniteness. Discussing lexical disambiguation, however, is
the simplest way to explain how underspecified representations can be given a
semantics related to Pinkal's proposals about ambiguity, and how to defeasible
reasoning with underspecified representations. In the next sections I will
generalize the approach introduced here to cases of ambiguity for which the
underspecified approach is much more plausible.  Furthermore, at least one
theory of lexical disambiguation, Hirst's proposal
\SHORTCITE{hirst:ambiguity-book}, makes use of `Polaroid words' which are
essentially underspecified interpretations of lexical items.

\subsection{A Lexically Underspecified Grammar}

The presentation of a lexically underspecified grammar below is centered on the
example of (H-type) lexical ambiguity discussed above, the verb
\LEXITEM{croak}, which can take two precisifications. Let $\cal L$ be the
language which consists of the single sentence \SENTENCE{Kermit croaked}. This
sentence is H-type ambiguous because it admits of two precisifications and it
is subject to the precisification imperative. A `Montagovian' grammar MG would
map (syntactic analyses of) the sentence into distinct expressions of a
`disambiguated language' ${\cal DL}$, each of which denotes a function from
discourse situations into intensional objects of the appropriate type (in this
case, propositions). A grammar UHG that subscribes to the Underspecification
Hypothesis, on the other hand, maps syntactic analyses of expressions of $\cal
L$ into a single expression of a `lexically underspecified language' ${\cal
LXUL}$. The semantics of ${\cal LXUL}$ is based on the Precisification
Principle: expressions of ${\cal LXUL}$ denote at each discourse situation a
{\em set} of senses of the type they would be assigned by a Montagovian
grammar.\footnote{The technique of assigning sets of senses as the denotation
of sentences dates back at least to Hamblin \SHORTCITE{hamblin:questions},
who used it to extend Montague's fragment to questions.}

The lexically underspecified language ${\cal LXUL}$ has the following
ingredients:

\begin{description}
  \item [Terms]: the constant \TERM{k}.
  \item [Predicates:] \PRED{croak$_U$},
        \PRED{croak$_1$}, and   \PRED{croak$_2$}.
  \item [Atomic Formulas:] If t is a term and P is a predicate, then
  	P(t) is a formula.
  \item [Formulas:]
  	If {\PHI} and {\PSI} are formulas, then {\NOT\PHI} and {\PHI} \AND
        {\PSI} are formulas.
\end{description}

Note that in addition to two predicates \PRED{croak$_1$} and \PRED{croak$_2$},
corresponding to the disambiguated senses of \LEXITEM{croak}, the language
includes an `underspecified' predicate \PRED{croak$_U$}.  The interpretation
function for ${\cal LXUL}$, \SHORTVALUE{\ALPHA}, is defined as follows.  Let M
= \PAIR{U}{F} be a model just like the one that would be used for a
disambiguated language ${\cal DL}$.  The interpretation function
\SHORTVALUE{\ALPHA} assigns to an expression {\ALPHA} of ${\cal LXUL}$ a value
with respect to M and a discourse situation d.

\begin{itemize}
  \item \SHORTVALUE{\TERM{k}} = \{f$_k$\}, where
  			         f$_k$ : s \RIGHTARROW the denotation of
  				  `Kermit' in s
  \item \SHORTVALUE{\PRED{croak$_1$}} = \{f$_1$\}, where
  				  f$_1$ is a function such
                                  that f$_1$(s) \RIGHTARROW
  				  the set of croaking objects in s
  \item \SHORTVALUE{\PRED{croak$_2$}} = \{f$_2$\}, where
  				  f$_2$ is a function such
                                  that f$_2$(s) \RIGHTARROW
  				  the set of dying objects in s\}
  \item \SHORTVALUE{\PRED{croak$_U$}} = \{f$_1$, f$_2$\}
  \item \SHORTVALUE{P(t)} = \{f $|$ for a g \IN \SHORTVALUE{P} and
                                    an h \IN  \SHORTVALUE{t}
                                    f(s) = g(s)[h(s)]
                                    \}
  \item \SHORTVALUE{{\PHI} \AND {\PSI}} =
                     \{f $|$        for a g \IN \SHORTVALUE{P} and
                                    an h \IN  \SHORTVALUE{t},
                                    f(s) = 0 if g(s) = 0 or
  				    h(s) = 0;
                                    f(s) = 1  if both g(s) = 1 and h(s) = 1
                                    \}
  \item \SHORTVALUE{\NOT{\PHI}} =
                     \{f $|$        for
                                    g \IN \SHORTVALUE{\PHI},
                     	            f(s) = 1 if g(s) = 0;
                                    f(s) = 0 if g(s) = 1 \}
\end{itemize}
The language ${\cal LXUL}$ has been deliberately kept  simple
to make it clear that the underspecified languages I propose have two basic
properties: (i) the value of an expression at a discourse situation is a set of
senses of the type that a sense of that expression would have in a
disambiguated language ${\cal DL}$; and (ii) expressions can be divided into
expressions whose denotation at a discourse situation is a singleton set, such
as \TERM{k} or \PRED{croak$_1$}, and expressions such as \PRED{croak$_U$} that
denote a non-singleton set.  The latter expressions provide the interpretation
for ambiguous expressions of $\cal L$.\footnote{A lexical item can also be
ambiguous in that it may be associated with lexical entries of different
syntactic categories: for example, the word \LEXITEM{duck} can either be
interpreted as a noun or as an verb, as shown by the example \SENTENCE{They saw
her duck}. I assume a syntactic ambiguity in these cases, i.e., I assume that
the grammar would assign two syntactic analyses to the word \LEXITEM{duck},
each of which would then get an interpretation in ${\cal LXUL}$.}

The clauses for application and the connectives show how ambiguity `percolates
up' from lexical items. The value of an expression like \LOGEXP{\ALPHA(\BETA)}
is obtained by taking the cross-product of the values of {\ALPHA} and {\BETA},
and it includes one function f per distinct pair of functions
\PAIR{\ALPHA$_1$}{\BETA$_1$} in the denotations of {\ALPHA} and {\BETA}. The
value assigned by the function f to the situation \INMODEL{s} is defined by
applying a certain operation (in this case, application) to the values assigned
to \INMODEL{s} by the functions $\alpha_1$ and $\beta_1$. Thus, if both the
denotation of {\ALPHA} and the denotation of {\BETA} are singleton sets, the
denotation of \ALPHA(\BETA) is also a singleton set; otherwise, ambiguity
`multiplies,' as it where.\footnote{It is perhaps worth emphasizing a
difference between the semantics just sketched and virtually all other
approaches to underspecification I am aware of. In this proposal, the
underspecified language ${\cal LXUL}$ does not serve as a `meta-language' to be
given a semantics in terms of the values assigned to the expressions of a
`disambiguated language'; instead, it has a semantics of its own, defined
bottom-up much in the way the semantics of ${\cal DL}$ would be defined. In
other words, the approach just sketched does not rely on the assumption that a
`disambiguated language' can be defined, which, at the light of Pinkal's
treatment of indefiniteness, appears to be questionable. For example, for
Pinkal an expression is `purely vague' is no natural precisification exists.}
The same `multiplication' technique is also used to define the denotation of
connectives.\footnote{I will ignore in what follows the issue of partiality,
e.g., what happens when a conjunct has a value other than 0 or 1.}

The following grammar generates an underspecified representation of
\SENTENCE{Kermit croaked} by mapping the semantically ambiguous predicate
\LEXITEM{croaked} into an `ambiguous' predicate of ${\cal LXUL}$ as follows:

\begin{itemize}
  \item S  \RIGHTARROW NP VP; VP\PRIME(NP\PRIME)
  \item NP \RIGHTARROW Kermit; \TERM{k}
  \item VP \RIGHTARROW croaked; \PRED{croak$_U$}
\end{itemize}
The underspecified translation of \SENTENCE{Kermit croaked} in ${\cal LXUL}$,
\PRED{croak$_U$}(\TERM{k}), denotes a set of two propositions at a situation d:
the function that assigns 1 to a situation iff Kermit produced a frog-like
sound in that situation, and the function that assigns 1 to a situation iff
Kermit died in that situation.  This makes the sentence indefinite in Pinkal's
sense.\footnote{The denotation assigned to an indefinite sentence by the
grammar above is a simplification of the denotation that Pinkal would assign to
such a sentence in \cite{pinkal:lal}, which, in addition the set of senses
associated with a natural language expression, would also include a partial
order relation of precisification between them. Such an order relation plays an
important role in the meaning of vague sentences such as \SENTENCE{Kermit is
tall} is considered, which has distinct senses depending on the degree of
precision with which the discourse situation is specified, but a less important
one in the cases of `narrow sense' ambiguity with which I am concerned here.}
By contrast, an indeterminate sentence such as \SENTENCE{Kermit is the
Ruritanian secretary of state} would have a single sense at a given discourse
situation.

Pinkal's Precisification Imperative is an attempt at making more precise the
observation that human beings don't seem to have good intuitions concerning
what follows from a H-type ambiguous sentence.  Even when subjects are able to
pass judgments about what follows from an ambiguous sentence, it's arguable
that they do not give judgments concerning what follows from the underspecified
representation: rather, they first generate one interpretation, then decide
what follows from that.  The conclusion that I would be inclined to draw is
that a relation of semantic entailment capturing human intuitions can only be
defined, if at all, between expressions whose interpretation is not subject to
the Precisification Imperative. So, although it would be possible to define,
for example, a `strong' notion of entailment as what follows from all senses,
this definition would be rather artificial. For this reason I will not attempt
to define a notion of entailment between expressions of ${\cal LXUL}$; the
readers interested in the issue are referred to Pinkal's book and to the
discussion in van Deemter's dissertation \SHORTCITE{van-deemter:thesis}.

\SUBSECTION{Discourse Interpretation and Perceived Ambiguity}{disc_int_section}

\subsubsection{Discourse Interpretation and Defeasible Reasoning}

A theory of ambiguity processing solves the Combinatorial Explosion Puzzle if
it does not require that all distinct interpretations of a semantically
ambiguous sentence are actually generated. A grammar consistent with the
Underspecification Hypothesis such as the one just discussed moves us one step
towards that goal, since it only imposes the constraint that a single
underspecified interpretation be generated.

On the other hand, we can conclude from the discussion of deliberate ambiguity
and of the psychological work on ambiguity that a psychologically plausible
theory of ambiguity must also predict that more than one interpretation may
become available in a given context, although the number of such
interpretations will in general be much smaller than the number of possible
semantic interpretations.

As discussed above, the view of discourse interpretation that I am going to
take is the one typically found in the AI literature, according to which
disambiguation involves the generation of (possibly distinct) hypotheses in
parallel by means of defeasible inference. This perspective is found, for
example, in the work on abductive discourse interpretation by Hobbs and
colleagues \CITE{hobbs-et-al:abduction-TR}, in the work on Bayesian
disambiguation by, e.g., Charniak and his students \CITE{charniak&goldman:88}
and in the work on DICE and discourse interpretation by Asher, Lascarides, and
Oberlander \CITE{lascarides-et-al:ACL92}. Some of the formal models of
defeasible reasoning that can be used to formalize the situation in which
conflicting hypotheses are generated include Reiter's \NEWTERM{default logic}
\CITE{reiter:default-reasoning}, the abductive model
\CITE{hobbs-et-al:abduction-TR}, Bayesian Nets, and {\small DICE}
\cite{lascarides-et-al:ACL92,asher&morreau:91}.

\subsubsection{Lexical Disambiguation Using Defaults}

As a model of defeasible reasoning, I adopt Reiter's Default Logic.  In
default logic, the process that generates defeasible hypotheses is seen as the
computation of the \NEWTERM{extensions} of a \NEWTERM{default theory} (D,W)
where D is a set of default inference rules and W is a set of formulas. I will
formalize discourse interpretation as the process of generating the extensions
of the theory (DI,UF), where DI---the \NEWTERM{Discourse Interpretation
Principles}---are default inference rules, and UF is a set of expressions of an
underspecified language like ${\cal LXUL}$. Let us ignore for the moment the
fact that the formulas in UF are underspecified representations. The Discourse
Interpretation Principles formalize the defeasible inferences that take place
in discourse interpretation, such as disambiguating inferences. These rules are
operations that map a set of wffs that allow of a certain number of
interpretations into a new set of wffs with a more restricted number of
interpretations. An example of Discourse Interpretation Principle is the
following:

\begin{quote}
{\bf CROAK1-IF-FROG:} \\[1ex]
\begin{tabular}{l}
                \PRED{croak$_U$}(\TERM{x}) \AND
                	 \PRED{frog}(\TERM{x}) :
                \PRED{croak$_1$}(\TERM{x})\\
                \hline
                \PRED{croak$_1$}(\TERM{x})
                \end{tabular}
\end{quote}
This inference rule\footnote{CROAK1-IF-FROG is an {\em open} inference
rule. Such rules act like inference rule schemas.}  reads: if the set of wffs
UF includes the fact that the object \TERM{x} has the property \PRED{croak$_U$}
and the property \PRED{frog}, and if it is consistent to assume that the
interpretation \PRED{croak$_1$} of \PRED{croak$_U$} was intended, then the
inference rule CROAK1-IF-FROG produces a new set of wffs that includes the fact
\PRED{croak$_1$}(\TERM{x}). The application of CROAK1-IF-HUMAN-LIKE would be
blocked by the presence in UF of the wff \NOT\UCOND{croak$_1$}{k}. Using
Reiter's definition of extension in an `intuitive' fashion, we can see that the
default theory

\begin{center}
  (\{CROAK1-IF-FROG\},
   \{\PRED{croak$_U$}(\TERM{k}) \AND
     \UCOND{frog}{k}\})
\end{center}
has the  following (unique) extension:

\begin{center}
   \{\UCOND{croak$_U$}{k} \AND
     \UCOND{frog}{k},
     \UCOND{croak$_1$}{k}\}
\end{center}
The denotation of a set of wffs \{\PHI$_1$ \ldots \PHI$_n$\} will be defined as
the denotation of the conjunction \PHI$_1$ \AND \ldots \AND \PHI$_n$ of these
wffs. I also assume that the empty set of wffs denotes the function TRUE that
is true at every situation. With this definition, and under the assumption that
each `unambiguous' interpretation of the word \LEXITEM{croak} is incompatible
with the others (i.e., under the assumption that \UCOND{croak$_1$}{x} $\vdash$
\NOT\UCOND{croak$_2$}{x}), the extension of (DF,UI) admits of only one
denotation, the one under which the denotation of \TERM{k} produced a sound
like the one frogs produce.\footnote{See the discussion below.}

A default theory always has an extension as long as all defaults are
normal,\footnote{I.e., of the form \ALPHA:\BETA / \BETA.} but it may have more
than one extension if the set of Discourse Interpretation Principles contains
two inference rules that both apply but generate a conflict. Consider, for
example, the default theory consisting of a set of discourse interpretation
principles {DI\PRIME} that includes, in addition to CROAK1-IF-FROG, a second
discourse interpretation principle (let's call it CROAK2-IF-HUMAN-LIKE) stating
that the \PRED{croak$_2$} interpretation is plausible for human-like beings;
and of a set of wffs {UF\PRIME} including the fact that Kermit is a human-like
being.

\begin{center}
  (\{CROAK1-IF-FROG,CROAK2-IF-HUMAN-LIKE\},\\
   \{\PRED{croak$_U$}(\TERM{k}) \AND
     \UCOND{frog}{k},
     \UCOND{human-like}{k}
     \})
\end{center}
this theory would have two extensions:

\begin{enumerate}
\item
   \{\UCOND{croak$_U$}{k} \AND
     \UCOND{frog}{k},
     \UCOND{human-like}{k},
     \UCOND{croak$_1$}{k}\}
\item
   \{\UCOND{croak$_U$}{k} \AND
     \UCOND{frog}{k},
     \UCOND{human-like}{k},
     \UCOND{croak$_2$}{k}\}
\end{enumerate}

Perceived ambiguity can now be redefined more precisely as the state that
obtains when the default theory `encoding' the listener's discourse
interpretation processes has more than one extension; and the cases of
deliberate ambiguity discussed in section \SECREF{ambiguity_section} can be
formalized as cases in which the speaker has `reasoned about the other agent's
reasoning,' as it were.

\subsubsection{Constraints on Discourse Interpretation and the
	     Anti-Random Hypothesis}

Once we start allowing discourse interpretation processes like those just
discussed, the Underspecification Hypothesis is not sufficient to explain the
Combinatorial Explosion Puzzle anymore. The UH does not rule out a theory of
discourse interpretation in which after an underspecified interpretation has
been obtained, all possible senses of a sentence are generated. In fact, a lot
of {\NLP} systems work this way, as well as interpretation procedures such as
Hobbs and Shieber's scoping algorithm \CITE{hobbs&shieber:87}. In the
framework for discourse interpretation just presented, theories of this kind
could be formalized by including discourse interpretation principles that
generate all the semantically justified interpretations at random. For the case
of lexical disambiguation, for example, we could have a theory that includes
the two following inference rules:

\begin{quote}
{\bf CROAK1-AT-RANDOM:} \\[1ex]
\begin{tabular}{l}
                \PRED{croak$_U$}(\TERM{x}):
                   \PRED{croak$_1$}(\TERM{x})\\
                \hline
                \PRED{croak$_1$}(\TERM{x})
                \end{tabular}\\[2ex]
\end{quote}
\TROUT{\newpage}
\begin{quote}
{\bf CROAK2-AT-RANDOM:} \\[1ex]
\begin{tabular}{l}
                \PRED{croak$_U$}(\TERM{x}):
                    \PRED{croak$_2$}(\TERM{x})\\
                \hline
                \PRED{croak$_2$}(\TERM{x})
\end{tabular}
\end{quote}

A theory of lexical disambiguation of this kind would simply produce all
semantically justified interpretations of a sentence, and the Combinatorial
Explosion Puzzle would remain a puzzle. To solve the puzzle, a theory of
disambiguation must therefore supplement the Underspecification Hypothesis with
constraints on discourse interpretation that ensure that only a few extensions
are generated.

The constraints need not be the same for all classes of ambiguity. For certain
classes of ambiguity, including perhaps lexical ambiguity,\footnote{See however
\cite{hirst:ambiguity-book}.}  the explanation may simply be that the
disambiguation process is {\em incremental}, i.e., it takes place as the text
is processed word by word or constituent by constituent, and each ambiguity is
resolved {\em locally}; in this way, only a small number of alternative
hypotheses have to be considered every time. For other classes of ambiguity,
however, such as scopal ambiguity and referential ambiguity, incremental
processing does not seem to be the solution,\footnote{In a sentence such as
\SENTENCE{John gave a present to each child}, for example, the indefinite
\LEXITEM{a present} takes narrow scope with respect to the quantifier
\LEXITEM{each child}. The interpretation of the sentence must therefore either
remain partially underspecified until the quantifier is processed, or be
revised when the quantifier is encountered. Similarly, when processing a
sentence such as \SENTENCE{John always invites MARY to the movies}, whose
preferred interpretation is that whenever John goes to the movies, he invites
Mary, the restriction of the adverb of quantification \SENTENCE{always} is not
encountered until the PP \LEXITEM{to the movies} is encountered.

\TRIN{ There is little doubt that part of the solution to the Combinatorial
Explosion Puzzle is that some forms of ambiguity, at least, are solved locally
and incrementally. Garden-path phenomena, for example, are commonly interpreted
as providing evidence for this hypothesis
{}~\cite{frazier&fodor:78,crain&steedman:85,altmann:parsing-interpretation},
and,
as discussed above, similar effects can exploit forms of ambiguity other than
syntactic ambiguity: e.g., scopal ambiguity. The examples just discussed
suggest however that an incremental account of discourse interpretation, as
well, must be supplemented with a theory of underspecification.} } and
different constraints must apply. In \cite{poesio:thesis}, the following
constraint was proposed:

\begin{description}
  \item [Anti-Random Hypothesis (Informal)]
Humans do not randomly generate alternative interpretations of an ambiguous
sentence; only those few interpretations are obtained that (i) are consistent
with syntactic and semantic constraints and (ii) are suggested by the context.
\end{description}
The Anti-Random Hypothesis should be thought of as a `meta-constraint' on
theories of interpretation: if we intend to account for the Combinatorial
Explosion Puzzle, we have to develop theories of interpretation (e.g., theories
of parsing, or theories of definite description interpretation) that satisfy
this constraint, i.e., in which discourse interpretation principles like
CROAK1-AT-RANDOM and CROAK2-AT-RANDOM are not allowed.

\KEESIN{ In order to illustrate more concretely the difference between theories
of discourse interpretation that satisfy the Anti-Random Hypothesis, and
theories that do not, let us consider how one could formalize a theory of
pronominal interpretation. A `random' theory of pronoun interpretation would go
as follows: first, compute all possible antecedents of the pronoun in the
discourse. Then, generate an hypothesis for each of them, stating that the
pronoun refers to that antecedent. Finally, rank these hypotheses according to
their plausibility. A random hypothesis generation process usually leaves the
task of choosing one hypothesis to plan recognition; the problem is that most
often, the alternatives are equally plausible.

In contrast, \NEWTERM{centering theory} \CITE{GJW:83} is an example of
non-random pronoun interpretation theory.  According to centering theory, each
utterance establishes a `backward looking center' (Cb), and a pronoun is by
default interpreted to refer to the Cb. (I am glossing over a number of
complexities here.)  Such a theory would generate a single (or a few)
hypothesis concerning the antecedent of a pronoun; the other possibilities,
although semantically possible, would simply never come up. Examples of
theories of definite description interpretation, tense interpretation, the
interpretation of modals in discourse, and scope disambiguation that satisfy
the Anti-Random Hypothesis are discussed in \cite{poesio:thesis}.

}

The Anti-Random Hypothesis can be made more formal in the framework for
discourse interpretation adopted here by introducing a slightly different
syntax for default inference rules, one in which the underspecified condition
is syntactically separated from additional contextual requirements such as the
requirement in CROAK1-IF-FROG that the object in question be a frog:

\begin{quote}
  \DIP{{\bf CROAK1-IF-FROG}}{
       \UCOND{croak$_U$}{x}}{
           \UCOND{frog}{x}}{
       \UCOND{croak$_1$}{x}}
\end{quote}
Except for the fact that one of the prerequisite wffs is `singled out', an
inference rule thus rewritten has the same interpretation as one of Reiter's
default rules. We can then require the contextual requirements to be
non-trivial (i.e., not satisfied in every situation) as follows:

\begin{description}
  \item [Anti-Random Hypothesis]
A discourse interpretation theory (DI,UF) is {\em Anti-Random} iff
for all  discourse interpretation principles {\ALPHA:\BETA:\DELTA / \GAMMA} in
DI, {\BETA} is not satisfied in every situation.
\end{description}

\subsubsection{The Condition on Discourse Interpretation}

The framework just introduced can also be used to formalize the `second order'
aspects of Pinkal's theory, such as the Precisification Imperative. The
Precisification Imperative can be seen as imposing a constraint on the
extensions of a discourse interpretation theory, namely, as the requirement
that extensions include a `disambiguating wff' like \UCOND{croak$_1$}{k} for
each H-type ambiguous constituent of the set UF such as \UCOND{croak$_U$}{k}. I
will call this constraint \NEWTERM{Condition on Discourse
Interpretation}. In first instance, the Condition on Discourse
Interpretation might be formulated as follows, for the case of lexical
ambiguity:

\begin{description}
  \item [Condition on Discourse Interpretation (Preliminary):] Each\\
extension E of a discourse interpretation theory (DI,UF) must include, for each
literal L in UF whose predicate is H-type ambiguous, a distinct
\NEWTERM{disambiguating literal}, i.e., a literal whose denotation is a single
function among those in the denotation of L.
\end{description}
The definition of the Condition on Discourse Interpretation just given is not
very general: it depends on the assumption that all cases of H-type ambiguity
are originated by predicates. A simpler, and more general, formulation of the
Condition on Discourse Interpretation can be obtained by generalizing the
format for the discourse interpretation principles once more.

Default inference rules are typically used to augment a set of wffs with
additional facts inferred by default: the fact that a particular bird flies,
for example. But the purpose of discourse interpretation rules used for
disambiguation, like CROAK1-IF-FROG, is to restrict the interpretation by
eliminating certain readings. In this perspective, leaving the underspecified
wffs around doesn't make much sense. I propose therefore to allow discourse
interpretation principles to {\em rewrite} their `triggering wff' whenever this
wff encodes an H-type ambiguity, in addition to adding new wffs to a set. The
more general format for discourse interpretation principles is as follows:

\begin{center}
  \DIPR{\ }{\ALPHA}{\BETA}{\DELTA}{\SIGMA}{\TAU}
\end{center}
A  rule of this form is an operation from sets of wffs into sets of wffs that
given a set W of wffs containing {\ALPHA} and {\BETA} and not containing
\NOT\DELTA,\footnote{Strictly speaking, one should check that {\NOT\DELTA} does
not occur in the extension itself, not in the intermediate sets of wffs; this
form makes sense however once we adopt a `syntactic' definition of extension
(see below).} produces a set {W\PRIME} of wffs containing {\TAU}, and in which
{\ALPHA} has been replaced by {\SIGMA}. I will call {\ALPHA} the
\NEWTERM{triggering condition}. For example, a version of CROAK1-IF-FROG in
which the triggering condition \UCOND{croak$_U$}{x} is rewritten by the
consequent \UCOND{croak$_1$}{x} is as follows:

\begin{center}
  \DIPR{{\bf CROAK1-IF-FROG}}{
       \UCOND{croak$_U$}{x}}{
           \UCOND{frog}{x}
             }{\UCOND{croak$_1$}{x}}{
       \UCOND{croak$_1$}{x}}{}
\end{center}
If all disambiguation rules are rewritten in this format,
a completely disambiguated extension can simply be characterized as one which
doesn't contain any \NEWTERM{H-type ambiguous wffs}. The notion of H-type
ambiguous wff can be characterized either syntactically (by identifying certain
syntactic constituents as specifying H-type ambiguity, and by classifying as
H-type ambiguous a wff that contains one of these constituents)\footnote{Both
the treatment of scopal ambiguity and the treatment of referential ambiguity
proposed below are such that the constituents that introduce H-type ambiguity
can be identified syntactically.} or model-theoretically, e.g., by means of a
function $\iota$ such that if X is a set of senses, $\iota$(X) is 1 if the set
of senses is admissible, 0 if it is inadmissible in Pinkal's sense.  Whatever
way we choose to define a H-type ambiguous wff, the Condition on Discourse
Interpretation can now be formulated as follows:

\begin{description}
  \item [Condition on Discourse Interpretation]: An
extension E of a discourse interpretation theory (DI,UF) cannot contain an
H-type ambiguous wff.
\end{description}
Notice that the statement of the Condition on Discourse Interpretation as a
condition on pragmatic reasoning gives it the status of a felicity condition
rather than of a hard constraint on interpretation.

\subsubsection{Extensions, closure and consistency checking in an
underspecified default theory}

So far, I've been using the terminology from default logic as if the shift to
an underspecified representation had no side effects, but this is not the
case. Consider the way in which Reiter defines the notion of extension of a
(closed)\footnote{A closed default theory is one in which no default contains
open variables. All really interesting cases of default inference rules do
include such variables; but Reiter derives the definition of the extension of
an `open' default theory from the definition of extension for closed theories.}
default theory, for example:

\begin{DEFINITION}
Let $\Delta$ = (D,W) be a closed default theory, so that every default of D has
the form (\ALPHA:\BETA$_1$,\ldots, \BETA$_m$/w) where
{\ALPHA},\BETA$_1$,\ldots,\BETA$_m$,w are all closed wffs of L. For any set of
closed wffs S \SUBSETEQ L let $\Gamma$(S) be the smallest set satisfying the
following three properties:
\begin{itemize}
  \item [D1] W \SUBSETEQ $\Gamma$(S)
  \item [D2] Th($\Gamma$(S)) = $\Gamma$(S)
  \item [D3] If (\ALPHA:\BETA$_1$,\ldots, \BETA$_m$/w) \IN D
  	and \ALPHA \IN S and \NOT\BETA$_1$, \ldots, \NOT\BETA$_m$ $\neg\in$
        S then w \IN $\Gamma$(S).
\end{itemize}
A set of closed wffs E \SUBSETEQ L is an \NEWTERM{extension} for $\Delta$
iff $\Gamma$(E) = E, i.e., iff E is a fixed point of the operator $\Gamma$.
\end{DEFINITION}
This definition crucially relies on the notion of deductive closure Th(S),
defined as the set of wffs \{w $|$ S $\vdash$ w\}; but what we said with
semantic entailment holds for provability, as well: no clear notion exists of
what it means for an expression of an underspecified language to follow from a
set of wffs of the same language.  Two routes are open to us. One is to define
a notion of `underspecified provability' $\vdash_U$, and to use $\vdash_U$ to
define an `underspecified' notion of closure Th$_U$(S). For example, we could
say that w $\vdash_U$ {w\PRIME} iff for each expression {w\PRIME\PRIME} that
denotes a single one of the interpretations of w, {w\PRIME\PRIME} $\vdash_U$
{w\PRIME}.  This route is not very appealing, however, if for no other reason
that it's not clear that any way of defining an underspecified notion of
provability will do.

The alternative is to adopt a new notion of extension that does not rely on
deductive closure, i.e., one in which an extension is a fixed point of the
operator $\Gamma'$, which does not include condition D2 of the definition of
$\Gamma$:

\begin{itemize}
  \item Let $\Gamma'$(S) be the smallest set satisfying the
following two properties:
\begin{itemize}
  \item [D1] W \SUBSETEQ $\Gamma'$(S)
  \item [D3] If (\ALPHA:\BETA$_1$,\ldots, \BETA$_m$/w) \IN D
  	and \ALPHA \IN S and \NOT\BETA$_1$, \ldots, \NOT\BETA$_m$ $\neg\in$
        S then w \IN $\Gamma'$(S).
\end{itemize}
\end{itemize}
Replacing $\Gamma$ with $\Gamma'$ in the definition of extension has several
consequences. First and foremost, dropping the requirement of deductive closure
makes the test of whether it is consistent to assume \BETA$_1$, \ldots,
\BETA$_m$ essentially syntactic: i.e., it is possible for \BETA$_j$ not to be
included in $\Gamma'$(S) even though it is derivable from $\Gamma'$(S). (In
general, this definition of extension is a much closer description of the
behavior of actual implementations of non-monotonic reasoning than the original
definition.) And therefore, a logic defined in this way does not have the
property of Reiter's logic that a (closed) default theory (D,W) has an
inconsistent extension iff W is inconsistent.

In fact, each extension of a discourse interpretation theory under this new
definition will, in general, be H-type ambiguous, some of the interpretations
being inconsistent. However, if we adopt the `rewriting' version of
disambiguation discussed above, and impose the Condition on Discourse
Interpretation, each extension will have a single interpretation, and therefore
its consistency can be checked. I propose therefore to define the notion of
extension of a discourse interpretation theory as follows:

\begin{description}
  \item [Extension:] A set of closed wffs E \SUBSETEQ L is an
\NEWTERM{extension} for the discourse interpretation theory $\Delta$
iff E is a fixed point of the operator $\Gamma'$ and satisfies the Condition on
Discourse Interpretation.
\end{description}

\section{Other Forms of Ambiguity}

The theory of ambiguity introduced in the previous sections can be
straightforwardly extended to obtain a treatment of two other classes of
semantic ambiguity: scopal ambiguity and referential ambiguity. These
extensions preserve the basic ideas of the theory, semantic ambiguity as
multiplicity of meanings, and perceived ambiguity as multiple extensions of a
default theory; what changes is that on the one hand, a more complex
underspecified language is introduced, capable of encoding other forms of
ambiguity; on the other hand, more complex inference rules are used.

\subsection{Scopal Ambiguity}

I will call the sentence constituents that modify the parameters of evaluation,
and therefore affect the interpretation of other sentence constituents `in
their scope', \NEWTERM{operators}. \TRIN{Examples of operators are quantifiers
(that affect the choice of the variable assignment used to evaluate expressions
in their scope) and modals (that affect the choice of the world / situation at
which expressions in their scope are evaluated).} As it is well-known, one
cause of semantic ambiguity is that sentences may contain more than one
operator, and their relative scope is not completely determined by the
sentence's syntactic structure.  Sentences that have more than one meaning due
to the interaction between operators are called \NEWTERM{scopally
ambiguous}.\footnote{`Ambiguity elimination' solutions to the combinatorial
explosion puzzle, such as Kempson and Cormack's \SHORTCITE{kempson&cormack:81}
or Verkuyl's \SHORTCITE{verkuyl:92} have had some success in showing that
certain cases of `ambiguity'---especially `ambiguities' associated with plural
noun phrases or certain classes of scopal ambiguities---are in fact cases of
indeterminacy.  Zwicky and Sadock noted that the identity tests do not classify
a sentence as ambiguous if the propositions expressed by the sentence are such
that one entails the other. This is the case, for instance, with sentences such
as \SREF{every-kid}
\cite{lakoff:vagueness-ambiguity,zwicky&sadock:75,kempson&cormack:81}.

\begin{EXAMPLE}
\ENEW{every-kid} Every kid climbed a tree.
\end{EXAMPLE}
\cite{kempson&cormack:81} claim that
sentences like \SREF{kc:1} are not ambiguous, but indeterminate: according to
them, such sentences semantically denote the weaker reading (the one in which
the universal quantifier takes scope over the existential).  The stronger
reading is the result of pragmatic reasoning.
\begin{EXAMPLE}
\ENEW{kc:1}  Every linguistics student has read a book by Chomsky.
\end{EXAMPLE}
Kempson and Cormack propose in fact that all quantified sentences denote a
single proposition; in this way, the combinatorial explosion puzzle
disappears, at least as far as scopal ambiguity is concerned. However, it is
not true in general that a sentence with two quantifiers has two
interpretations, one of which entails the other.  \SREF{weak:counter} does not
have an interpretation weak enough to be entailed by all others, yet able to
capture the truth conditions correctly.

\begin{EXAMPLE}
\ENEW{weak:counter} Few students know many languages.
\end{EXAMPLE}
A second  problem with the proposal  of Kempson and Cormack is that if one
wants to claim that the meaning of a sentences such as \SREFA{kmcd:2}{a} is
something like \SREFA{kmcd:2}{b}, as Kempson and Cormack do, then one ends up
predicting that the meaning of \SREFA{kmcd:2}{c} should be something like
\SREFA{kmcd:2}{d}, the strongest interpretation of the sentence. In other
words, one either has to give up compositionality for sentences like
\SREFA{kmcd:2}{b}, or to abandon the strategy of letting sentences
denote their weakest interpretation \CITE{chierchia&mcconnell:book}.

\begin{EXAMPLE}
\ENUMA{kmcd:2}
\EITEM Every kid climbed a tree.
\EITEM (\FORALL x \PRED{kid}(x) \SUPSET (\EXISTS y \PRED{tree}(y) \AND
                                                \PRED{climb}(x,y)))
\EITEM It is not the case that every kid climbed a tree.
\EITEM \NOT(\FORALL x \PRED{kid}(x) \SUPSET (\EXISTS y \PRED{tree}(y) \AND
                                                \PRED{climb}(x,y)))
\ENDENUMA
\end{EXAMPLE}
It should also be clear that whatever the case for scopal ambiguity, other
kinds of ambiguity, such as structural and H-type lexical ambiguity, cannot be
reduced to indeterminacy. }

Historically, most underspecified representations have been introduced to deal
with scopal ambiguity.  Typically, an intermediate step of processing is
assumed in which operators are left `in place,' as well as a subsequent step of
processing in which their relative scope is determined by contextual
processing. Schubert and Pelletier's underspecified representation of
\SENTENCE{Every kid climbed a tree} in \SREF{kmcd:3:ULF} is an example of
underspecified representation in which the operators are left `in situ'.

These representations are typically justified in terms of ease of processing,
and their ability to represent `intermediate' readings. It is clear however
that for the purposes of developing a `principled' theory of ambiguity
processing, it would be much better to stick to as few new `levels of
representation' as possible.

In fact, there is no need to introduce a new level of representation. The two
requirements on a scopally underspecified representation---that it allow
representing the structural information provided by a sentence, and
representing the intermediate steps of disambiguation---can be satisfied by
using as an underspecified representation the syntactic structure of the
sentence, augmented with information about the semantic interpretation of
word-forms. In this way we can also maintain semantic translation of lexical
items used in Montague grammar, that determine how they combine with other
sentence constituents to determine a sentence's meaning.

The `lexically and scopally underspecified language' ${\cal LSUL}$ I introduce
to encode scopal ambiguity generalizes the language ${\cal LXUL}$ introduced in
the previous section by allowing for arbitrary functional types. In this way,
the lexical item \LEXITEM{every} can be given its usual
\FTYPE{\FTYPE{e}{t}}{\FTYPE{\FTYPE{e}{t}}{t}} translation:
\begin{center}
 \LAMBDAEXP{P}{
   \LAMBDAEXP{Q}{\FORALLEXP{x}{\TERM{P}(\TERM{x})}{\TERM{Q}(\TERM{x})}}}.
\end{center}
The second augmentation to ${\cal LXUL}$ is the inclusion of tree-like
expressions used to translate syntactic phrases. For example, the {\NP}
\PSBRACKET{NP}{\PSBRACKET{Det}{every} \PSBRACKET{N}{dog}} translates into the
expression:

\begin{center}
  \PSBRACKET{NP}{
    \PSBRACKET{Det}{
      \LAMBDAEXP{P}{\LAMBDAEXP{Q}{
        \FORALLEXP{x}{\TERM{P}(\TERM{x})}{\TERM{Q}(\TERM{x})}}}}
    \PSBRACKET{N}{\PRED{dog}}}
\end{center}
The
expression in \SREF{undersp:ex} is the underspecified translation of
the sentence \SENTENCE{Every dog saw a frog}:

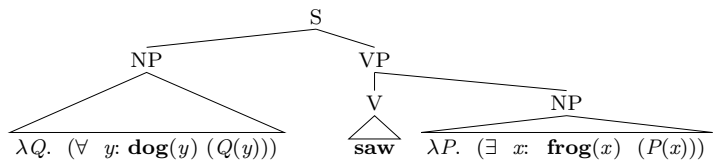
\begin{EXAMPLE}
\ENEW{undersp:ex} \ \ \\
     {\footnotesize
         \begin{tabular}{cccccc}
                        & \node{s}{S} \\[2ex]
         \node{npa}{NP} & & \node{vpa}{VP} \\[2ex]
                        & & \node{vc}{V} &  \node{npb}{NP} \\[2ex]
         \node{I}{\LAMBDAEXP{Q}{
                  \FORALLEXP{y}{\UCOND{dog}{y}}{\TERM{Q}(\TERM{y})}}}
                        & & \node{vsem}{\PRED{saw}}
                        & \node{npbsem}{\LAMBDAEXP{P}{
                                         \EXISTSEXP{x}{
                                            \UCOND{frog}{x}
                                            }{\TERM{P}(\TERM{x})}}}
        \end{tabular}
        \nodeconnect{s}{npa}\nodeconnect{s}{vpa}
        \nodeconnect{vpa}{vc}\nodeconnect{vpa}{npb}
        \nodetriangle{npa}{I}
        \nodetriangle{vc}{vsem}
        \nodetriangle{npb}{npbsem}
        }
\end{EXAMPLE}
Besides reducing the
number of representations floating around, this was of talking about scopal
underspecification has two additional advantages over underspecified
representations in which all syntactic information except for the position of
operators is lost, such as Schubert and Pelletier's underspecified logical
forms, Reyle's underspecified {\DRS}s or the Core Language Engine's
{\QLF}s. First of all, the semantics of expressions such as \SREF{undersp:ex}
---that, for historical reasons, I call \NEWTERM{logical forms}--- can be
computed in a completely classical fashion using the storage mechanism
\CITE{cooper:83}, with the result that syntactic constraints on the available
readings, such as the Scope Constraint \CITE{may:85,heim:82}, can play a role
in determining the semantics of these objects, without the need for additional
constraints such as the label ordering constraints used in UDRS
\CITE{reyle:93}. Secondly, all structural information is preserved, not just
information about the relative position of operators. Some of this syntactic
information is used as a clue during disambiguation, for example, for
interpreting pronouns, but also in certain theories of scopal
disambiguation. (See, e.g., \cite{kurtzman&macdonald:93} and
\cite{poesio:thesis} for an account of scope disambiguation which esploits the
syntactic information encoded by underspecified expressions such as
\SREF{undersp:ex}.)

\subsubsection{A Lexically and Scopally Underspecified Language}

The semantics of the underspecified language  ${\cal
LSUL}$ is classically based on a set {\TYPES} of semantic types, the smallest
set such that (i) \TYPE{e} and \TYPE{t} are types; and (ii) If
\TYPE{\TAU} and \TYPE{\TAU\PRIME} are types, \FTYPE{\TAU}{\TAU\PRIME} is a
type.  The set of meaningful expressions of type {\ALPHA} is indicated by
ME$_{\alpha}$. The set of non-logical constant expressions of type {\ALPHA} is
indicated as CE$_{\alpha}$ \SUBSETEQ ME$_{\alpha}$.

The semantics of ${\cal LSUL}$ is based on the same idea as the semantics of
${\cal LXUL}$. Natural language expressions are assigned objects of the same
type that they would receive in \cite{dowty-et-al:81} (as revised by Partee and
Rooth \shortcite{partee&rooth:83}), with the difference that, when I talk about
`meaningful expressions of type {\TAU}' below, therefore, I am really talking
about expressions that denote sets of functions from the set of situations
{\SITUATIONS} to elements of \DOMAIN{\TAU} (the domain of type {\TAU}). Thus
for example, sentences are of type \TYPE{t} both in Dowty, Wall and Peters'
system, and in the current proposal; but a meaningful expression of type
\TYPE{t} in ${\cal LSUL}$ denotes a (function from a discourse situation to a)
set of functions from situations to truth values. \KEESOUT{Or to make another
example, relations have the same type \FTYPE{e}{\FTYPE{e}{t}} here that they
have in Dowty, Wall and Peters, but meaningful expressions of type
\FTYPE{e}{\FTYPE{e}{t}} now denote sets of functions from {\SITUATIONS} to
\DOMAIN{\FTYPE{e}{\FTYPE{e}{t}}}.}

The sets of meaningful expressions of ${\cal LSUL}$ include all the
expressions in ${\cal LXUL}$:
\begin{itemize}
  \item CE$_{e}$ = \{\TERM{k}\}.
  \item CE$_{\FTYPE{e}{t}}$ =
\{\PRED{dog},\PRED{frog},\PRED{croak$_U$}, \PRED{croak$_1$}, \PRED{croak$_2$}
	\}.\\
        (The language includes a single
underspecified lexical interpretation, \PRED{croak$_U$}.)
  \item CE$_{\FTYPE{e}{\FTYPE{e}{t}}}$ = \{\PRED{saw}\}.
  \item If {\ALPHA} \IN ME$_{\FTYPE{\TAU}{\TAU\PRIME}}$ and {\BETA} \IN
  	ME$_{\TAU}$, \ALPHA(\BETA) \IN ME$_{\TAU\PRIME}$.
  \item If {\ALPHA} and {\BETA} are of type \TYPE{t}, then
  	{\ALPHA} \AND {\BETA} and {\NOT\ALPHA} are in ME$_t$.
\end{itemize}
The set ME$_{\TAU}$, for any type {\TAU}, includes a denumerably infinite
set of \NEWTERM{variables} of type \TYPE{\TAU}. ${\cal LSUL}$ also includes
lambda-abstracts and quantified expressions, defined below. The language also
includes the new syntactic category of \NEWTERM{logical forms}. The sets of
logical forms of syntactic category XP, LF$_{XP}$, are defined as follows:

\begin{itemize}
  \item LF$_{\mbox{Det}}$ =
  	\{\PSBRACKET{Det}{\ALPHA} $|$  {\ALPHA} \IN
        ME$_{\FTYPE{\FTYPE{e}{t}}{\FTYPE{\FTYPE{e}{t}}{t}}}$\};
  \item LF$_{\mbox{N}}$ =
  	\{\PSBRACKET{N}{\ALPHA} $|$ {\ALPHA} \IN
        ME$_{\FTYPE{e}{t}}$\};
  \item LF$_{\mbox{PN}}$ =
  	\{\PSBRACKET{PN}{\ALPHA} $|$ {\ALPHA} \IN  CE$_{\TYPE{e}}$\}
  \item LF$_{\mbox{IV}}$ =
  	\{ \PSBRACKET{IV}{\ALPHA} $|$ {\ALPHA} \IN
        ME$_{\FTYPE{e}{t}}$\};
  \item LF$_{\mbox{TV}}$ =
  	\{\PSBRACKET{TV}{\ALPHA} $|$ {\ALPHA} \IN
        ME$_{\FTYPE{e}{\FTYPE{e}{t}}}$\};
  \item LF$_{\mbox{NP}}$ =
  	\{\PSBRACKET{NP}{{\ALPHA} {\BETA}} $|$
          {\ALPHA} \IN LF$_{\mbox{Det}}$ and {\BETA} \IN
          LF$_{\mbox{N}}$ \} \\
          \UNION
        \{\PSBRACKET{NP}{{\ALPHA}} $|$
          {\ALPHA} \IN LF$_{\mbox{PN}}$ \} \UNION
        \{\PSBRACKET{NP}{{\ALPHA}} $|$
          {\ALPHA} a variable of type \TYPE{e} \}
  \item LF$_{\mbox{VP}}$ =
  	\{\PSBRACKET{VP}{{\ALPHA} {\BETA}} $|$
          {\ALPHA} \IN LF$_{\mbox{TV}}$ and {\BETA} \IN
          LF$_{\mbox{NP}}$ \} \\
          \UNION
        \{\PSBRACKET{VP}{{\ALPHA}} $|$
          {\ALPHA} \IN LF$_{\mbox{IV}}$ \}
  \item LF$_{\mbox{S}}$ =
  	\{\PSBRACKET{S}{{\ALPHA} {\BETA}} $|$
          {\ALPHA} \IN LF$_{\mbox{NP}}$ and {\BETA} \IN
          LF$_{\mbox{VP}}$ \}
\end{itemize}
Meaningful expressions are assigned a value with respect to a universe
{\UNIVERSE}.  I use below the notation \INMODEL{s} to indicate that \INMODEL{a}
`stands for' an object in {\UNIVERSE}, i.e., it is part of the metalanguage, as
opposed to being a meaningful expression of the object language. The models
with respect to which a {\CRT} expression is evaluated include a set
{\SITUATIONS} of situations. The only fact about situations I use here is that
they have \NEWTERM{constituents}.

The interpretation of types with respect to {\UNIVERSE} is defined as usual:
D$_{e,U}$ = \UNIVERSE; D$_{t,U}$ = \{0,1\}; D$_{\langle a,b \rangle,U}$ =
D$_b^{D_a}$. In the rest of this paper, I generally drop the indication of the
universe (e.g., I write D$_{e}$ instead of D$_{e,U}$).  The model of
interpretation for {\CRT} expressions is the triple \LANGLE \UNIVERSE,
\SITUATIONS, I\RANGLE. The interpretation function `I' assigns an
interpretation to constants of type {\TAU}.

The value of meaningful expressions is specified by a function \MGDVALUE{.}
that includes an assignment function among its parameters, since the terms of
${\cal LSUL}$ include variables. The interpretation of variables is specified
by the following clause:

\begin{itemize}
  \item For {\ALPHA} a variable of type \TYPE{\TAU},
        \MGDVALUE{\ALPHA}  = the
        singleton set \{g(\ALPHA)\}, where g(\ALPHA) is a constant function
        f :  {\SITUATIONS} \RIGHTARROW \DOMAIN{\TAU}.

\end{itemize}
The interpretation of constants, connectives and application is as in ${\cal
LXUL}$. The denotation of the other expressions is discussed below.

\subsubsection{An Example of a Scopally and Lexically Underspecified Grammar}

The following grammar extends the grammar discussed in section
\SECREF{lexamb_section} by adding determiners and relations as new lexical
items:
\begin{itemize}
  \item PN  \RIGHTARROW Kermit;  \TERM{k}
  \item Det \RIGHTARROW every;   \LAMBDAEXP{P}{
  				  \LAMBDAEXP{Q}{
                                    \FORALLEXP{x}{\TERM{P}(\TERM{x})
                                                }{\TERM{Q}(\TERM{x})}}}
  \item Det \RIGHTARROW a;       \LAMBDAEXP{P}{
  				  \LAMBDAEXP{Q}{
                                    \EXISTSEXP{x}{\TERM{P}(\TERM{x})
                                                }{\TERM{Q}(\TERM{x})}}}
  \item N   \RIGHTARROW dog;     \PRED{dog}
  \item N   \RIGHTARROW frog;    \PRED{frog}
  \item IV  \RIGHTARROW croaked; \PRED{croak$_U$}
  \item TV  \RIGHTARROW saw;     \PRED{saw}
\end{itemize}
and by adding  phrase structure rules for {\NP}s and transitive verbs:

\begin{itemize}
  \item S  \RIGHTARROW NP VP;   \PSBRACKET{S}{{NP\PRIME} {VP\PRIME}}
  \item NP \RIGHTARROW PN;      \PSBRACKET{NP}{\PSBRACKET{PN}{PN\PRIME}}
  \item NP \RIGHTARROW Det N;   \PSBRACKET{NP}{\PSBRACKET{Det}{Det\PRIME}
                 			       \PSBRACKET{N}{N\PRIME}}
  \item VP \RIGHTARROW IV;      \PSBRACKET{VP}{\PSBRACKET{IV}{V\PRIME}}
  \item VP \RIGHTARROW TV NP;   \PSBRACKET{VP}{\PSBRACKET{TV}{V\PRIME}
                                               {NP\PRIME}}
\end{itemize}
This grammar generates, in addition to lexically ambiguous sentences such as
\SENTENCE{Kermit croaked}, scopally ambiguous sentences such as \SENTENCE{Every
dog saw a frog}.

\subsubsection{The Denotation of Logical Forms}

The denotation of logical forms is specified using the {\em storage} method,
developed by Robin Cooper \INA{Cooper, R.}  as a way around a problem with
Montague's \NEWTERM{quantifying in} technique, namely, the fact that in order
to get all the readings of a scopally ambiguous sentence, one has to stipulate
that the sentence is syntactically ambiguous (see \cite{dowty-et-al:81}).

Cooper proposed that the value of a syntactic tree is a set of
\NEWTERM{sequences}, each sequence representing a distinct `order of
application' of the operators that may result in a admissible interpretation of
a sentence. For example, the quantifier \LEXITEM{a frog} can `enter' the
derivation of the {\VP} \LEXITEM{saw a frog} in two different ways. The narrow
scope reading is obtained by immediately applying the interpretation of the
quantifier to the translation of \LEXITEM{saw}; but it is also possible to
apply the predicate to the variable quantified over, and `wait' before applying
the quantifier, in which case the wide scope reading is obtained. The value of
the {\NP} \LEXITEM{a frog}, then, is the set of two sequences shown in
\SREF{cooper-storage:NP}. One sequence consists of a single element, the
`traditional' Montague-style translation of \LEXITEM{every frog}. The second
sequence consists of two elements: the variable \TERM{y}, and the semantic
translation of the quantified {\NP}, put `in storage'.

\begin{EXAMPLE}
\ENEW{cooper-storage:NP}
\begin{tabbing}
\{\=\SEQUENCE{\LAMBDAEXP{P}{
              \QEXP{\EXISTS}{y}{\UCOND{frog}{y}}{\TERM{P}(\TERM{y})}}},\\
 \>\SEQUENCE{\TERM{y},
           \LAMBDAEXP{P}{
              \QEXP{\EXISTS}{y}{\UCOND{frog}{y}}{\TERM{P}(\TERM{y})}}}
\}
\end{tabbing}
\end{EXAMPLE}
Ambiguity `propagates up' as follows. The value of the {\VP} \LEXITEM{saw a
frog} in \SREF{csex-a} also consists of two sequences, one obtained by applying
the first element of the first sequence in the denotation of \LEXITEM{every
frog} to the predicate \PRED{saw}, the other obtained by applying the predicate
\PRED{saw} to the first element of the second sequence (the variable
\TERM{y}). The result is as in \SREF{cooper-storage:VP}.

\begin{EXAMPLE}
\ENEW{csex-a}
 \PSBRACKET{VP}{\PSBRACKET{V}{\PRED{saw}}
                \PSBRACKET{NP}{\LAMBDAEXP{P}{
                                \QEXP{\EXISTS}{y}{\UCOND{frog}{y}
                                               }{\TERM{P}(\TERM{y})}}}}
\ENEW{cooper-storage:VP}
\begin{tabbing}
\{\=\LANGLE\LAMBDAEXP{x}{
             \QEXP{\EXISTS}{y}{\UCOND{frog}{y}}{\BCONDCURRY{saw}{y}{x}}}
    \RANGLE,\\
 \>\LANGLE \UCOND{saw}{y},
           \LAMBDAEXP{P}{
               \QEXP{\EXISTS}{y}{\UCOND{frog}{y}}{\TERM{P}(\TERM{y})}}
   \RANGLE
\}
\end{tabbing}
\end{EXAMPLE}
Finally, the value  of a sentence is obtained by combining the value
of the {\VP} with the value of the {\NP} in the usual fashion: the value of
\PSBRACKET{S}{\SENTENCE{Every dog saw a frog}} is a set of two sequences, each
representing a distinct reading of the sentence.

It's easy to see that Cooper's technique can be used to assign to
underspecified representations like \SREF{undersp:ex} a `multiple sense'
denotation like those assigned to lexically ambiguous expressions in the
previous section. All that is needed is a function CV that assigns to each
expression of the form \PSBRACKET{XP}{\ALPHA} its `Cooper Value'; the
denotation of sentence translations like \PSBRACKET{S}{\BETA} can then be
defined in terms of CV as follows:

\begin{itemize}
  \item If \PSBRACKET{S}{\ALPHA} is a logical form in LE$_{\mbox{S}}$
  	then \PSBRACKET{S}{\ALPHA} is a meaningful
        expression of type \TYPE{t}.
        Let CV(\ALPHA)(M,g,d) be the set
  	of single-element sequences
        \{\LANGLE\{$\sigma_{1_1}$,\ldots,$\sigma_{1_j}\}$\RANGLE,
          \ldots,
          \LANGLE\{$\sigma_{n_1}$,\ldots,$\sigma_{n_k}$\}\RANGLE\}. Then \\
        \MGDVALUE{\PSBRACKET{S}{\ALPHA}}
                      = \{$\sigma_{1_1}$,\ldots,$\sigma_{1_j}$,\ldots
                           $\sigma_{n_1}$,\ldots,$\sigma_{n_k}$\}
\end{itemize}
(I have taken into account the fact that an expression of our
underspecified language denotes a set of objects, therefore each scopally
disambiguated translation of a sentence will still denote a set of
propositions.)

Cooper discusses in detail in \cite{cooper:83} how semantic and syntactic
constraints on scope can be implemented as requirements that the storage be
`discharged' at certain positions---\IE, that no element in storage be `carried
across' syntactic constructions that produce scope islands, such as \BAR{S}. In
this way, no operator in a clause may take scope over operators in an higher
clause, or in a sister clause\TRIN{, thus enforcing the Scope Constraint
discussed in \SECREF{syn_sem_constr_section}}.\TRIN{\footnote{The definition of
storage generates spurious readings in the case of embedded NPs such as
\LEXITEM{a representative of every company}, that have to be eliminated via a
separated filter \CITE{keller:nested-cooper-storage}. Keller introduced a
`nesting' technique that obviates the problem
\shortcite{keller:nested-cooper-storage}.  More recently, Pereira
\cite{pereira:CL90} argued that the right scoping properties can be obtained
without additional stipulations from the natural deduction approach to parsing.
I only consider here `basic' NPs that do not create problems for the simplest
version of Cooper's technique.}}

The CV function used to define the interpretation of logical forms is based on
an implementation of the storage idea less general than Cooper's, but simpler.
In order to arrive at a uniform specification of the Cooper Value of all
constructs, it is useful to define construct-specific versions of application
in which to `bury' the differences in storage manipulation. These operations
are defined as follows:

\begin{itemize}
  \item \TDAPPLY{\ALPHA}{\BETA}{S} =
          \FUNDEF{\item \ALPHA(\BETA),
                        if Type(\ALPHA) = \FTYPE{\FTYPE{e}{t}}{t}
                        and Type(\BETA) = \FTYPE{e}{t}, \\
                        or if
                        Type(\ALPHA) = \FTYPE{\FTYPE{e}{t}}{\FTYPE{e}{t}}
                        and Type(\BETA) = \\
                        \FTYPE{e}{t}.
                  \item \TDAPPLY{\LAMBDAEXP{y}{
                                  \LAMBDAEXP{x}{
                                    \BETA(\TERM{x})(\TERM{y})}}}{\ALPHA}{VP}
                  	if Type(\ALPHA) = \FTYPE{\FTYPE{e}{t}}{t} and \\
                        Type(\BETA) = \FTYPE{e}{\FTYPE{e}{t}}, or if
                        Type(\ALPHA) = \\
                        \FTYPE{\FTYPE{e}{t}}{\FTYPE{e}{t}}
                        and Type(\BETA) = \FTYPE{e}{\FTYPE{e}{t}}.
                  \item \BETA(\ALPHA),
                        if Type(\BETA) = \FTYPE{e}{t}
                        and Type(\ALPHA) = \TYPE{e}.
                  \item Undefined otherwise.}
  \item \TDAPPLY{\ALPHA}{\BETA}{VP} =
          \FUNDEF{\item \ALPHA(\BETA),
                        if Type(\ALPHA) = \FTYPE{e}{t}
                        and Type(\BETA) = \TYPE{e}.
                  \item \LAMBDAEXP{w$_e$}{
                          \BETA(\ALPHA(\TERM{w$_e$}\/))},
                        if Type(\BETA) = \FTYPE{\FTYPE{e}{t}}{\FTYPE{e}{t}}\\
                        and Type(\ALPHA) = \FTYPE{e}{\FTYPE{e}{t}}.
                  \item \LAMBDAEXP{w$_e$}{
                          \BETA(\LAMBDAEXP{z}{
                                    \ALPHA(\TERM{z}\/)(\TERM{w$_e$}\/)})},
                        if Type(\BETA) = \FTYPE{\FTYPE{e}{t}}{t}\\
                        and Type(\ALPHA) = \FTYPE{e}{\FTYPE{e}{t}}.
                  \item Undefined otherwise.}
  \item \TDAPPLY{\ALPHA}{\BETA}{XP} (where XP is a
  	category other than S or VP) = \\
          \FUNDEF{\item \ALPHA(\BETA),
                        if Type(\ALPHA) = \FTYPE{\TAU}{\TAU\PRIME}
                        and Type(\BETA) = \TYPE{\TAU}.
                  \item \BETA(\ALPHA),
                        if Type(\BETA) = \FTYPE{\TAU}{\TAU\PRIME}
                        and Type(\ALPHA) = \TYPE{\TAU}.
                  \item Undefined otherwise.}
\end{itemize}
Next, we need an operation that
combines two sets of sequences into one. The result of applying this operation
to two sets of sequences X and Y is the set of sequences obtained by (typed)
applying the first element of a sequence in X to the first element of a
sequence in Y and then merging the rest of the sequences, as follows:

\begin{itemize}
  \item
  \STORECOMBINE{X}{Y}{YP}, where X and Y are the sets \\
  X = \{\SEQUENCE{x$_{11}$,x$_{12}$, \ldots x$_{1m}$}, \ldots
        \SEQUENCE{x$_{n1}$,x$_{n2}$, \ldots x$_{nl}$}\} and \\
  Y = \{\SEQUENCE{y$_{11}$,y$_{12}$, \ldots y$_{1p}$}, \ldots
        \SEQUENCE{y$_{q1}$,y$_{q2}$, \ldots y$_{qr}$}\}, \\
  and YP is any phrase category, is the set: \\
  \{
    \SEQUENCE{\TDAPPLY{x$_{i1}$}{y$_{j1}$}{YP},
              y$_{j2}$, \ldots y$_{jp}$}, x$_{i2}$, \ldots x$_{im}$ $|$
              \SEQUENCE{x$_{i1}$,x$_{i2}$, \ldots x$_{im}$} \IN X and
              \SEQUENCE{y$_{j1}$,y$_{j2}$, \ldots y$_{jp}$} \IN Y
   \}
\end{itemize}
We also need an operation to put operators into store, and one to
`discharge' them. The {\STOREAWAY} operation takes a set consisting a single
single-element sequence and a \NEWTERM{result}, and returns a set that consists
of two sequences: the original sequence, and a new sequence consisting of the
result and the operator in store.

\begin{itemize}
  \item \STOREAWAY(\ALPHA,X), where X = \{\SEQUENCE{\BETA}\}, is
  	\{\SEQUENCE{\BETA},\SEQUENCE{\ALPHA,\BETA}\}
\end{itemize}
The {\DISCHARGE} operation takes a sequence and applies all operators back to
obtain a set of sequences with a single element and an empty store. For
simplicity, we will assume that all operators are generalised quantifiers,
i.e., of type \FTYPE{\FTYPE{e}{t}}{t}. (No other operators are specified in the
grammar above.) {\DISCHARGE} is defined as follows:

\begin{itemize}
  \item \DISCHARGE(\SEQUENCE{x$_{1}$,x$_{2}$, \ldots x$_{m}$}) =
  	\{\TDAPPLY{x$_{m}$}{
            \TDAPPLY{x$_{m-1}$}{\ldots
              \TDAPPLY{x$_{2}$}{x$_{1}$}{S}}{S}
              }{S} \}
\end{itemize}

\DISCHARGE$^*$(X), where X is a set of sequences, is the union $\bigcup_{x \in
X}$ \DISCHARGE($x$). We can now specify the Cooper value of logical forms with
respect to model M, variable assignment g, and discourse situation d as
follows:\footnote{Strictly speaking, the form CV(\ALPHA)(M,g,d) should be
used. I omit the indices below.}

\begin{itemize}
  \item CV(\PSBRACKET{S}{\PSBRACKET{NP}{\ALPHA} \PSBRACKET{VP}{\BETA}}) =
          \DISCHARGE$^*$(\STORECOMBINE{CV(\PSBRACKET{NP}{\ALPHA})
                          }{CV(\PSBRACKET{VP}{\BETA})}{S})
  \item CV(\PSBRACKET{NP}{\PSBRACKET{PN}{\ALPHA}}) =
  	  CV(\PSBRACKET{PN}{\ALPHA})
  \item CV(\PSBRACKET{NP}{\PSBRACKET{Det}{\ALPHA}
                          \PSBRACKET{N}{\BETA}}) =
          \STOREAWAY(\LAMBDAEXP{Q}{\LAMBDAEXP{z}{\TERM{Q}(\TERM{z})}},
             \STORECOMBINE{CV(\PSBRACKET{Det}{\ALPHA})
                      }{CV(\PSBRACKET{N}{\BETA})}{NP})
  \item CV(\PSBRACKET{NP}{\ALPHA}) =
  	  \{\SEQUENCE{\MGDVALUE{\ALPHA}}\}, where {\ALPHA} is a variable
          of type \TYPE{e}.
  \item CV(\PSBRACKET{VP}{\PSBRACKET{IV}{\ALPHA}}) =
          CV(\PSBRACKET{IV}{\ALPHA})
  \item CV(\PSBRACKET{VP}{\PSBRACKET{TV}{\ALPHA} \PSBRACKET{NP}{\BETA}}) =
             \STORECOMBINE{CV(\PSBRACKET{TV}{\ALPHA})
                         }{CV(\PSBRACKET{NP}{\BETA})}{VP}
  \item CV(\PSBRACKET{PN}{PN\PRIME}) = \{\SEQUENCE{\MGDVALUE{PN\PRIME}}\}
  \item CV(\PSBRACKET{Det}{Det\PRIME}) = \{\SEQUENCE{\MGDVALUE{Det\PRIME}}\}
  \item CV(\PSBRACKET{IV}{IV\PRIME}) = \{\SEQUENCE{\MGDVALUE{IV\PRIME}}\}
  \item CV(\PSBRACKET{TV}{TV\PRIME}) = \{\SEQUENCE{\MGDVALUE{TV\PRIME}}\}
  \item CV(\PSBRACKET{N}{N\PRIME}) = \{\SEQUENCE{\MGDVALUE{N\PRIME}}\}
\end{itemize}
There are three tricky aspects to  the definition of CV: the discharge
operation in the definition of the Cooper Value of a sentence translation, the
definition of CV(\PSBRACKET{NP}{\PSBRACKET{Det}{\ALPHA} \PSBRACKET{N}{\BETA}})
in which an operator is put in store, and the definition of
CV(\PSBRACKET{VP}{\PSBRACKET{V}{\ALPHA} \PSBRACKET{NP}{\BETA}} in which two
stores are combined, and that has different results depending on whether the
{\NP} is of type \TYPE{e} or is a quantifier. I'll illustrate these cases by
looking at the main steps of the computation of the CV of \SREF{undersp:ex}:

\begin{enumerate}
  \item CV(\PSBRACKET{Det}{
            \LAMBDAEXP{P}{
              \LAMBDAEXP{Q}{
             \QEXP{\EXISTS}{y}{\TERM{P}(\TERM{y})}{\TERM{Q}(\TERM{y})}}}}) =\\
  \{\SEQUENCE{\LAMBDAEXP{P}{
               \LAMBDAEXP{Q}{
                \QEXP{\EXISTS}{y}{\TERM{P}(\TERM{y})}{\TERM{Q}(\TERM{y})}}}}\}
  \\
  (In what follows, I'll use \PRED{a} to indicate
   the lexical translation of the determiner \LEXITEM{a}.)
  \item CV(\PSBRACKET{NP}{\PSBRACKET{Det}{\PRED{a}}
                          \PSBRACKET{N}{\PRED{frog}}}) = \\
    \STOREAWAY(\LAMBDAEXP{Q}{\LAMBDAEXP{z}{\TERM{Q}(\TERM{z})}},
       \STORECOMBINE{CV(\PSBRACKET{Det}{\PRED{a}})
                }{CV(\PSBRACKET{N}{frog})}{NP}) = \\
  \{\SEQUENCE{\LAMBDAEXP{Q}{
                \QEXP{\EXISTS}{y}{\UCOND{frog}{y}}{\TERM{Q}(\TERM{y})}}},\\
    \SEQUENCE{\LAMBDAEXP{Q}{\LAMBDAEXP{z}{\TERM{Q}(\TERM{z})}},
    	      \LAMBDAEXP{Q}{
                \QEXP{\EXISTS}{y}{\UCOND{frog}{y}}{\TERM{Q}(\TERM{y})}}}
  \}\\
  (I'll use {[\PRED{a frog}]} to stand for
  {\PRIME\LAMBDAEXP{Q}{
                \QEXP{\EXISTS}{y}{\UCOND{frog}{y}}{\TERM{Q}(\TERM{y})}}\PRIME}
   below.)
  \item CV(\PSBRACKET{VP}{\PSBRACKET{TV}{\PRED{saw}}
  		  \PSBRACKET{NP}{\PSBRACKET{Det}{\PRED{a}}
                                 \PSBRACKET{N}{\PRED{frog}}}}) = \\
             \STORECOMBINE{CV(\PSBRACKET{TV}{\PRED{saw}})
                         }{CV(\PSBRACKET{NP}{\PSBRACKET{Det}{\PRED{a}}
                               \PSBRACKET{N}{\PRED{frog}}})}{VP} =\\
        \{\SEQUENCE{\TDAPPLY{\PRED{saw}}{[\PRED{a frog}]}{VP}},
          \SEQUENCE{\TDAPPLY{\PRED{saw}}{
                             \LAMBDAEXP{Q}{\LAMBDAEXP{z}{\TERM{Q}(\TERM{z})}}
                            }{VP},
                    [\PRED{a frog}]}
        \} =\\
        \{\SEQUENCE{\LAMBDAEXP{w}{
                     \QEXP{\EXISTS}{y}{\UCOND{frog}{y}
                        }{\BCONDCURRY{saw}{y}{w}}}},\\
          \SEQUENCE{\LAMBDAEXP{w}{\LAMBDAEXP{x}{\BCONDCURRY{saw}{w}{x}}},
                    [\PRED{a frog}]}
        \} \\
        (Notice that the first element of the second sequence,
         in which [\PRED{a frog}] is kept in store, is still of type
         \FTYPE{e}{\FTYPE{e}{t}}.) I will use the abbreviations
         {\SIGMA} and {\SIGMA\PRIME} below  for the first element of the
         first sequence and the first element of the second sequence,
         respectively.
  \item CV(\PSBRACKET{NP}{\PSBRACKET{Det}{\PRED{every}}
                          \PSBRACKET{N}{\PRED{dog}}}) = \\
  \{\SEQUENCE{\LAMBDAEXP{Q}{
                \QEXP{\FORALL}{x}{\UCOND{dog}{x}}{\TERM{Q}(\TERM{x})}}},\\
    \SEQUENCE{\LAMBDAEXP{Q}{\LAMBDAEXP{z}{\TERM{Q}(\TERM{z})}},
    	      \LAMBDAEXP{Q}{
                \QEXP{\FORALL}{x}{\UCOND{dog}{x}}{\TERM{Q}(\TERM{x})}}}
  \}\\
  (I'll use {[\PRED{every dog}]} to stand for
  {\PRIME\LAMBDAEXP{Q}{
                \QEXP{\FORALL}{x}{\UCOND{dog}{x}}{\TERM{Q}(\TERM{x})}}\PRIME}
   below.)
  \item CV(\PSBRACKET{S}{\PSBRACKET{NP}{\PSBRACKET{Det}{\PRED{every}}
                                        \PSBRACKET{N}{\PRED{dog}}}
                         \PSBRACKET{VP}{\PSBRACKET{V}{\PRED{saw}}
  		  \PSBRACKET{NP}{\PSBRACKET{Det}{\PRED{a}}
                                 \PSBRACKET{N}{\PRED{frog}}}}}) =
          \DISCHARGE$^*$(\STORECOMBINE{CV(\PSBRACKET{NP}{\ALPHA})
                          }{CV(\PSBRACKET{VP}{\BETA})}{S}) = \\
\DISCHARGE$^*$\(\left(\mbox{\begin{tabular}{l}
                \{\SEQUENCE{\TDAPPLY{[\PRED{every dog}]}{\SIGMA}{S}},\\
                 \SEQUENCE{\TDAPPLY{[\PRED{every dog}]}{\SIGMA\PRIME}{S},
                           [\PRED{a frog}]},\\
                 \SEQUENCE{\TDAPPLY{
                            \LAMBDAEXP{Q}{\LAMBDAEXP{z}{\TERM{Q}(\TERM{z})}}}{%
                            \SIGMA}{S},[\PRED{every dog}]},\\
                 \SEQUENCE{\TDAPPLY{
                            \LAMBDAEXP{Q}{\LAMBDAEXP{z}{\TERM{Q}(\TERM{z})}}}{%
                            \SIGMA\PRIME}{S},
                              [\PRED{every dog}], [\PRED{a frog}]}\}
                 \end{tabular}
                 }\right)\) = \\
\DISCHARGE$^*$\(\left(\mbox{\begin{tabular}{l}
                \{\SEQUENCE{
                  \QEXP{\FORALL}{x}{\UCOND{dog}{x}}{
                    \QEXP{\EXISTS}{y}{\UCOND{frog}{y}}{
                      \BCONDCURRY{saw}{y}{x}}}},\\
                 \SEQUENCE{\LAMBDAEXP{w}{
                     \QEXP{\FORALL}{x}{\UCOND{dog}{x}
                        }{\BCONDCURRY{saw}{w}{x}}},[\PRED{a frog}]},\\
                 \SEQUENCE{\LAMBDAEXP{z}{
                     \QEXP{\EXISTS}{y}{\UCOND{frog}{y}}{
                      \BCONDCURRY{saw}{y}{x}}},[\PRED{every dog}]},\\
                 \SEQUENCE{\LAMBDAEXP{w}{
                     \LAMBDAEXP{z}{\BCONDCURRY{saw}{z}{w}}},
                           [\PRED{a frog}],[\PRED{every dog}]}
                \}\end{tabular}
                 }\right)\) = \\
\{\SEQUENCE{
                  \QEXP{\FORALL}{x}{\UCOND{dog}{x}}{
                    \QEXP{\EXISTS}{y}{\UCOND{frog}{y}}{
                      \BCONDCURRY{saw}{y}{x}}}},\\
  \SEQUENCE{
                  \QEXP{\EXISTS}{y}{\UCOND{frog}{y}}{
                      \QEXP{\FORALL}{x}{\UCOND{dog}{x}}{
                         \BCONDCURRY{saw}{y}{x}}}}\}
\end{enumerate}
The Scope Constraint is enforced by requiring a complete discharge
at the sentential level, which means no operators can `move up' outside the
sentence in which it occurs, although of course this couldn't occur in this
grammar since it doesn't cover relative clauses, sentential complements or
coordination.  I have assumed that discharge only takes place at sentential
level, i.e., there are no operators taking scope over {\VP}s; doing this would
complicate matters a bit in that a `partial' discharge operation should be
defined.\KEESOUT{\footnote{Some arguments for VP scope are discussed in
\cite{may:85,carpenter:scope}.}}

\subsubsection{Lambda Abstracts}

Some care is required in the system developed here to get a semantics for
lambda-abstraction that preserves properties such as \BETA- and
$\eta$-reduction.  The clause specifying the denotation of lambda-abstraction
in Dowty, Wall and Peters's book is the following:

\begin{itemize}
  \item If {\ALPHA} is a variable of type {\TAU} and {\BETA} a meaningful
        expression
        of type {\TAU\PRIME},
         then {\LAMBDA \ALPHA.\BETA} is an expression of type
        \FTYPE{\TAU}{\TAU\PRIME},
        and \STILVALUE{\LAMBDA \ALPHA.\BETA} is that
        function h: \DOMAIN{\TAU}  \RIGHTARROW \DOMAIN{\TAU\PRIME}
        such that for all objects \INMODEL{a} in \DOMAIN{\TAU},
        h(\INMODEL{a}) is equal to
        \ILVALUE{\BETA}{M}{g\{\ALPHA/\INMODEL{a}\}}{w}{t}.
\end{itemize}
If we generalize this clause in the `obvious'  way we get:

\begin{itemize}
  \item If {\ALPHA} is a variable of type {\TAU}
        and {\BETA} a  meaningful expression of type {\TAU\PRIME},
        then \LOGEXP{\LAMBDA \ALPHA.\BETA} is an expression of type
        \FTYPE{\TAU}{\TAU\PRIME},
        and \MGDVALUE{\LAMBDA \ALPHA.\BETA} is the
        set
        \{f $|$ where
        f: {\SITUATIONS}
        \RIGHTARROW (\DOMAIN{\TAU} \RIGHTARROW \DOMAIN{\TAU\PRIME})
        and
        for all situations \INMODEL{s},
        f(\INMODEL{s}) = h: \DOMAIN{\TAU} \RIGHTARROW \DOMAIN{\TAU\PRIME},
        such that, for all objects \INMODEL{a} in \DOMAIN{\TAU},
         h(\INMODEL{a}) is equal to h\PRIME(\INMODEL{s}),
        for some
        {h\PRIME} \IN
            \DENOTATION{\BETA}\mbox{$^{\mbox{M,g\{\ALPHA/\INMODEL{a}\},d}}$}
         \}.
\end{itemize}
Lambda-abstraction defined in this way does not have the required
properties. To show that it does not preserve $\eta$-reduction,\footnote{I.e.,
that \LAMBDAEXP{\ALPHA}{\BETA(\ALPHA)} \NEQ \BETA.}  it is sufficient to
consider the following example: let {\SITUATIONS} = \{\INMODEL{s$_1$},
\INMODEL{s$_2$}\}, \DOMAIN{\TAU} = \{\INMODEL{a},\INMODEL{b}\}, and let the
expression {\BETA} of type \FTYPE{\TAU}{\TAU\PRIME} have the following
denotation:

\begin{center}
  $\left\{ \mbox{
  \outerfs{\INMODEL{s$_1$} \RIGHTARROW
                     \outerfs{\INMODEL{a} \RIGHTARROW \INMODEL{$\alpha_1$}\\
                              \INMODEL{b} \RIGHTARROW
\INMODEL{$\beta_1$}}\\[2ex]
           \INMODEL{s$_2$} \RIGHTARROW
                     \outerfs{\INMODEL{a} \RIGHTARROW \INMODEL{$\alpha_2$}\\
                              \INMODEL{b} \RIGHTARROW \INMODEL{$\beta_2$}}},
  \outerfs{\INMODEL{s$_1$} \RIGHTARROW
                     \outerfs{\INMODEL{a} \RIGHTARROW \INMODEL{$\alpha_3$}\\
                              \INMODEL{b} \RIGHTARROW
\INMODEL{$\beta_3$}}\\[2ex]
           \INMODEL{s$_2$} \RIGHTARROW
                     \outerfs{\INMODEL{a} \RIGHTARROW \INMODEL{$\alpha_4$}\\
                              \INMODEL{b} \RIGHTARROW \INMODEL{$\beta_4$}}}
                }
  \right\}$
\end{center}
Then \VALUE{\LOGEXP{\BETA(\ALPHA)}}{M,g\{\ALPHA/\INMODEL{a}\},d} is as
follows:
\begin{center}
  $\left\{ \mbox{
  \outerfs{\INMODEL{s$_1$} \RIGHTARROW \INMODEL{$\alpha_1$}\\
           \INMODEL{s$_2$} \RIGHTARROW \INMODEL{$\alpha_2$}},
  \outerfs{\INMODEL{s$_1$} \RIGHTARROW \INMODEL{$\alpha_3$}\\
           \INMODEL{s$_2$} \RIGHTARROW \INMODEL{$\alpha_4$}}
                }
  \right\}$
\end{center}
and \VALUE{\LOGEXP{\BETA(\ALPHA)}}{M,g\{\ALPHA/\INMODEL{b}\},d}is as
follows:

\begin{center}
  $\left\{ \mbox{
  \outerfs{\INMODEL{s$_1$} \RIGHTARROW \INMODEL{$\beta_1$}\\
           \INMODEL{s$_2$} \RIGHTARROW \INMODEL{$\beta_2$}},
  \outerfs{\INMODEL{s$_1$} \RIGHTARROW \INMODEL{$\beta_3$}\\
           \INMODEL{s$_2$} \RIGHTARROW \INMODEL{$\beta_4$}}
                }
  \right\}$
\end{center}
Then, under the definition above, \MGDVALUE{\LAMBDAEXP{\ALPHA}{\BETA(\ALPHA)}}
will contain the following function, that is not part of the denotation of
{\BETA} (hence, $\eta$-reduction is not a sound inference rule):

\begin{center}
  \outerfs{\INMODEL{s$_1$} \RIGHTARROW
                   \outerfs{\INMODEL{a} \RIGHTARROW \INMODEL{$\alpha_1$}\\
                            \INMODEL{b} \RIGHTARROW \INMODEL{$\beta_3$}}\\[2ex]
           \INMODEL{s$_2$} \RIGHTARROW
                   \outerfs{\INMODEL{a} \RIGHTARROW \INMODEL{$\alpha_4$}\\
                            \INMODEL{b} \RIGHTARROW \INMODEL{$\beta_2$}}}
\end{center}
Intuitively, the problem with the definition above is that it does not
`preserve' the functions in the denotation of {\BETA}. A definition of
lambda-abstraction that does preserve these functions, and therefore preserves
the soundness of \BETA- and $\eta$-reduction, can be obtained as
follows.\footnote{I wish to thank an anonymous reviewer for suggesting this
solution to the problem just discussed, much simpler than the solution proposed
in \cite{poesio:thesis}.}

The denotation function \MGDVALUE{\ALPHA} used so far assigns a value to
expression {\ALPHA} \IN ME$_{\TAU}$ in model M with respect to the parameters
of evaluation g and d, \MGDVALUE{\ALPHA} \SUBSETEQ {\SITUATIONS} \RIGHTARROW
\DOMAIN{\TAU}. Another way of specifying the value of expressions is to define
a function \MDVALUE{\ALPHA} that assigns as value to {\ALPHA} at discourse
situation d a set of functions of type (Ass \RIGHTARROW (\SITUATIONS
\RIGHTARROW \DOMAIN{\TAU})), from assignments to functions in (\SITUATIONS
\RIGHTARROW \DOMAIN{\TAU}). For example, Dowty, Wall and Peters' clause for
lambda abstraction could be rewritten as follows:

\begin{itemize}
  \item If {\ALPHA} is a variable of type {\TAU} and {\BETA} a meaningful
        expression
        of type {\TAU\PRIME},
         then {\LAMBDA \ALPHA.\BETA} is an expression of type
        \FTYPE{\TAU}{\TAU\PRIME},
        and \DENOTATION{\LAMBDAEXP{\ALPHA}{\BETA}} =
        $\Lambda$(\ALPHA,\DENOTATION{\BETA}), where
        $\Lambda$(\ALPHA,Y) is that function f from Ass \RIGHTARROW
        (\DOMAIN{\TAU} \RIGHTARROW \DOMAIN{\TAU\PRIME})
        such that for all g \IN Ass, \INMODEL{s} \IN
        \SITUATIONS, \INMODEL{a} in \DOMAIN{\TAU},
        f(g)(\INMODEL{s})(\INMODEL{a}) =
               Y(g\{\ALPHA/\INMODEL{a}\})(\INMODEL{s}).
\end{itemize}
This definition can then be generalized as follows:

\begin{itemize}
  \item If {\ALPHA} is a variable of type {\TAU} and {\BETA} a meaningful
        expression
        of type {\TAU\PRIME},
         then \LAMBDAEXP{\ALPHA}{\BETA} is an expression of type
        \FTYPE{\TAU}{\TAU\PRIME},
        and \MDVALUE{\LAMBDAEXP{\ALPHA}{\BETA}} =
        $\Lambda^+$(\ALPHA,\DENOTATION{\BETA}), where
        $\Lambda^+$(\ALPHA,Y) = \{$\Lambda$(\ALPHA,m) for some m \IN Y\} =
        \{f \IN (Ass
           \RIGHTARROW
           (\SITUATIONS \RIGHTARROW
              (\DOMAIN{\TAU} \RIGHTARROW \DOMAIN{\TAU\PRIME})))
        such that for all g \IN Ass,
        \INMODEL{s} \IN \SITUATIONS, \INMODEL{a} in \DOMAIN{\TAU}, and for
        some m \IN Y,
        f(g)(\INMODEL{s})(\INMODEL{a}) =
               m(g\{\ALPHA/\INMODEL{a}\})(\INMODEL{s}). \}
\end{itemize}
Lambda-abstraction defined this way does support
$\eta$-reduction.\footnote{The proof is as
follows. \MDVALUE{\LAMBDAEXP{\ALPHA}{\BETA(\ALPHA)}} = \{f \IN (Ass \RIGHTARROW
(\SITUATIONS \RIGHTARROW (\DOMAIN{\TAU} \RIGHTARROW \DOMAIN{\TAU\PRIME}))) such
that for all g \IN Ass, \INMODEL{s} \IN \SITUATIONS, \INMODEL{a} in
\DOMAIN{\TAU}, and for some m \IN \VALUE{\BETA(\ALPHA)}{M},
f(g)(\INMODEL{s})(\INMODEL{a}) = m(g\{\ALPHA/\INMODEL{a}\})(\INMODEL{s}). \}
Because of the definition of \MDVALUE{\BETA(\ALPHA)}, this is the set of
functions f such that f(g)(\INMODEL{s})(\INMODEL{a}) =
p(g\{\ALPHA/\INMODEL{a}\})(\INMODEL{s})[q(g\{\ALPHA/\INMODEL{a}\})(\INMODEL{s})],
for some p \IN \VALUE{\BETA}{M} and some q in \VALUE{\ALPHA}{M}, i.e., of the
functions which occur in \VALUE{\BETA}{M} since
q(g\{\ALPHA/\INMODEL{a}\})(\INMODEL{s}) = \INMODEL{a}.}  Since this more
general way of assigning a value is not needed to provide a semantics for the
other constructs of ${\cal LSUL}$, I will continue using a function
\MGDVALUE{.}, but the reader should keep in mind that a denotation function of
this form is needed to deal with lambda abstraction, hence, with
quantification. (And for referential ambiguity, as we will see below.)

\subsubsection{Quantification}

The treatment of quantifiers in ${\cal LSUL}$ is based on Generalized
Quantifiers Theory \CITE{barwise&cooper:81}, i.e., the idea that determiners
denote relations between two sets. The `restricted quantification' notation
used in the examples above is defined in terms of the two determiners
\DET{every} and \DET{a}, as follows:

\begin{itemize}
  \item \QEXP{\FORALL}{x}{\PHI}{\PSI} $\equiv_{\mbox{def}}$
        \DET{every}(\LAMBDAEXP{x}{\PHI},\LAMBDAEXP{x}{\PSI})
  \item \QEXP{\EXISTS}{x}{\PHI}{\PSI} $\equiv_{\mbox{def}}$
        \DET{a}(\LAMBDAEXP{x}{\PHI},\LAMBDAEXP{x}{\PSI})
\end{itemize}
A `single-valued'  semantics for
\DET{every}(\LAMBDAEXP{\ALPHA}{\PHI},\LAMBDAEXP{\ALPHA}{\PSI}) could be
defined, in first approximation, as in the following clause:

\begin{itemize}
  \item  
         Let F,G be meaningful expressions of type \FTYPE{e}{t}.
         Then \DET{every}(F,G)
         is a meaningful expression of type \TYPE{t}, and
         \MGDVALUE{\DET{every}(F,G)}
         is  the function f s.t.

         f(\INMODEL{s}) = \FUNDEF{
                \item [undefined] iff either
                   \MGDVALUE{F}  or
                   \MGDVALUE{G} are  \\
                   undefined;
                \item [1] iff \{\INMODEL{a} \IN {\UNIVERSE} s.t.
                              [\MGDVALUE{F}(\INMODEL{s})](\INMODEL{a})
                               = 1 \} \SUBSETEQ \\
                              \{\INMODEL{b} \IN {\UNIVERSE} s.t.
                              {[}\MGDVALUE{G}(\INMODEL{s}){]}(\INMODEL{b})
                               = 1 \}.
                \item [0] otherwise.
                   }

\end{itemize}
This definition can be generalized as follows into one that
works in the case in which \MGDVALUE{.} is a set:

\begin{itemize}
  \item
\MGDVALUE{\DET{every}(F,G} =

  	\{f $|$ there is an h \IN \MGDVALUE{F}
                and an {h\PRIME} \IN \MGDVALUE{G} such that

                f(\INMODEL{s}) = \FUNDEF{
                \item [undefined] iff either   h(\INMODEL{s})
                   is undefined or h\PRIME(\INMODEL{s}) is undefined;
                \item [1] iff for every \INMODEL{a} s.t.
                   [h(\INMODEL{s})](\INMODEL{a}) = 1,
                   [h\PRIME(\INMODEL{s})](\INMODEL{a}) = 1.
                \item [0] otherwise.
                }

        \}
\end{itemize}
The interpretation of expressions of the form
\DET{a}(\LAMBDAEXP{\ALPHA}{\PHI},\LAMBDAEXP{\ALPHA}{\PSI}) is defined in a
similar fashion, with the obvious semantics.

\subsubsection{Scope Disambiguation by Defeasible Inference}

Having extended the language into one that can be used to describe scopal
underspecification, the framework for discourse interpretation developed in
section \SECREF{disc_int_section} can also be used to formalize the inferences
involved in scope disambiguation.  Partially disambiguated interpretations can
be represented by expressions which mix logical forms with `traditional'
expressions, as done in {\DRT}. For example, one could formalize Ioup's
\SHORTCITE{ioup:75} Grammatical Function Principle, stating that an {\NP} in
subject position by default takes scope over {\NP}s in other position, as
follows:

\begin{center}
{\footnotesize
  \DIPR{{\bf GRAMMATICAL-FUNCTION-PRINCIPLE}}
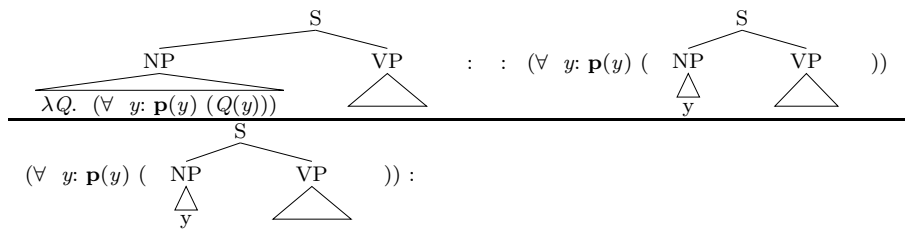
{
  	\begin{tabular}{cccccc}
                        & \node{s}{S} \\[2ex]
         \node{npa}{NP} & & \node{vpa}{VP} \\[2ex]
         \node{npasem}{\LAMBDAEXP{Q}{
                  \QEXP{\FORALL}{y}{\UCOND{p}{y}}{\TERM{Q}(\TERM{y})}}}
                        & & \node{vpasem}{\ \ \ \ \ \ \ \ \ }
        \end{tabular}
        \nodeconnect{s}{npa}\nodeconnect{s}{vpa}
        \nodetriangle{npa}{npasem}
        \nodetriangle{vpa}{vpasem}}{\ }{
        \QEXP{\FORALL}{y}{\UCOND{p}{y}}{
        \begin{tabular}{cccccc}
                        & \node{s}{S} \\[2ex]
         \node{npa}{NP} & & \node{vpa}{VP} \\[2ex]
         \node{npasem}{y}
                        & & \node{vpasem}{\ \ \ \ \ \ \ }
        \end{tabular}
        \nodeconnect{s}{npa}\nodeconnect{s}{vpa}
        \nodetriangle{npa}{npasem}
        \nodetriangle{vpa}{vpasem}}
        }{
        \QEXP{\FORALL}{y}{\UCOND{p}{y}}{
        \begin{tabular}{cccccc}
                        & \node{s}{S} \\[2ex]
         \node{npa}{NP} & & \node{vpa}{VP} \\[2ex]
         \node{npasem}{y}
                        & & \node{vpasem}{\ \ \ \ \ \ \ \ \ }
        \end{tabular}
        \nodeconnect{s}{npa}\nodeconnect{s}{vpa}
        \nodetriangle{npa}{npasem}
        \nodetriangle{vpa}{vpasem}}}{\ }
}
\end{center}
Logical forms in LF$_S$ are sentential expressions, and can therefore
serve as triggering condition of discourse interpretation principles. They can
also occur embedded in other expressions of ${\cal LSUL}$. During the scope
disambiguation process, `less ambiguous' expressions are inferred by deriving
expressions such as
\QEXP{\FORALL}{y}{\UCOND{p}{y}}{\PSBRACKET{S}{\PSBRACKET{NP}{y}
\PSBRACKET{VP}{\ALPHA}}} in which some quantifiers have been extracted, by a
process very similar to the one used in the top-down version of the {\DRT}
construction algorithm \cite{kamp&reyle:93}. `Partial' scopal disambiguation is
thus represented by ${\cal LSUL}$ expressions which still contain logical
forms.

As some readers will have already observed, the rule
GRAMMATICAL-FUNCTION-PRINCIPLE does not satisfy the Anti-Random restriction
proposed in \SECREF{disc_int_section}: the rule does not contain a non-trivial
restriction on the contexts in which it can operate. The already mentioned
proposal in \cite{poesio:thesis} overcomes this problem by making the
activation of scope disambiguation rules depend on whether the appropriate
domain for the quantifier (its \NEWTERM{resource situation}) has been
identified; a presentation of that proposal would however require introducing
too much additional material.

\subsection{Referential Ambiguity}

\subsubsection{Referential Expressions as Cases of Semantic Ambiguity}

Yet another way in which the semantics of sentences is `underspecified' by
their syntax is in the interpretation of anaphoric expressions and other
expressions whose interpretation has to be fixed in context. In semantics,
referential expressions are traditionally translated as free variables whose
interpretation depends on the choice of an assignment function (for the cases
of deictic anaphora) or by assigning them the same variable bound by the
quantifier that serves as their antecedent (for the cases of bound
anaphora). This translation does capture the intuition that the truth
conditions of a sentence containing a referential expression can only be
evaluated after fixing the value of the referential expressions. It is also
clear, however, that distinct propositions are obtained depending on the value
assigned to these expressions, much as distinct propositions are obtained
depending on the choice of an interpretation for lexical items, or of a scope
for operators: in other words, a sentence which includes a referential
expression is semantically ambiguous much in the way a sentence containing a
lexically ambiguous item is.\footnote{Pinkal takes pretty much the same
position in \cite{pinkal:lal}. He also introduces a distinction there between a
`speaker-oriented' perspective on meaning versus a `hearer-oriented'
perspective. A speaker may well have a single interpretation in mind for a
particular anaphoric expressions, but the hearer may have to recover this
interpretation among the many that are possible in that particular
context. This is of course true of all kinds of ambiguities, also those which
have a pragmatic rather than a semantic nature, but in the case of referential
ambiguity, the alternative interpretations correspond to distinct propositions
in the semantic sense as well. }

A complete discussion of reference interpretation would require introducing a
formalization of context, so I will only consider here the issue of providing
an underspecified treatment of intra-clausal and deictic anaphora. I propose
that referential expressions are cases of semantic ambiguity, and translate
into a special kind of underspecified object that I will call
\NEWTERM{parameters}.  Semantically, a parameter is a type \TYPE{e} expression
that, in a discourse situation \INMODEL{d}, denotes a set of functions from
situations to elements of \DOMAIN{e} in \INMODEL{d}. For example, the pronoun
\LEXITEM{he} would translate into a parameter \PAR{x} which, in a discourse
situation \INMODEL{d} with constituents \INMODEL{a$_1$} \ldots \INMODEL{a$_n$},
and given the set {\SITUATIONS} of situations, will denote a set of functions
\{f$_1$, \ldots, f$_m$, \ldots\} from situations in {\SITUATIONS} to
\INMODEL{a$_1$} \ldots \INMODEL{a$_n$}, including at least the set of all
constant functions that map each situation \INMODEL{s} into \INMODEL{a$_j$} if
\INMODEL{a$_j$} is a constituent of that situation (see below), and the set of
all variable denotations.  The reader will immediately realize that parameters
are the equivalent for type \TYPE{e} expressions of `underspecified predicates'
like \PRED{croak$_U$} introduced above.\footnote{The term `parameter' comes
from Situation Semantics (e.g., \cite{gawron&peters:90}), where the lexical
items whose interpretation depends on context are called \NEWTERM{parametric},
in the sense that their interpretation depends on the value assigned in context
to one or more parameters. Parameters are also used in situation theory to
translate pronouns and other anaphoric expression; but although the name and
the `dotted' notation is preserved here, the parameters I have just introduced
are an entirely different type of objects than the parameters of situation
theory, which are a special sort of objects in the universe, entirely distinct
from individuals.}

More formally, I propose to extend the set of terms of ${\cal LSUL}$ with a new
class of parameters, whose interpretation is defined as follows. First of all,
let us reformulate the semantics of variables given before, and make variables
functions from assignments to values (rather than the other way around). This
involves again using as interpretation function one that maps expressions into
functions from assignments to meanings, as done for lambda-abstracts.

\begin{itemize}
  \item For {\ALPHA} a variable of type \TYPE{\TAU},
        \MDVALUE{\ALPHA}  = the
        singleton set \{f\}, where f is a  function
        f :  Ass \RIGHTARROW ({\SITUATIONS} \RIGHTARROW \DOMAIN{\TAU})
        such that for every assignment g, f(g) is a constant function from
        {\SITUATIONS} \RIGHTARROW \DOMAIN{\TAU}.
\end{itemize}
This definition of the meaning of a variable allows us to abstract
away from assignments. We can now define the semantics of parameters as
follows:
\begin{itemize}
  \item For {\ALPHA} a parameter of type \TYPE{\TAU},
        \MGDVALUE{\ALPHA}  = $\displaystyle\bigcup_{\beta_{\tau}}$
        \MDVALUE{\BETA$_{\tau}$} (the union of the denotation of all variables
        of type \TYPE{\TAU}) $\bigcup$ \{f $|$
        f : \SITUATIONS\ \RIGHTARROW D$_\tau$
        is a function  such that for each \INMODEL{s},
        f(\INMODEL{s}) = \INMODEL{a},
        where \INMODEL{a}
        is an object of \TYPE{\TAU} that
        is a constituent of a situation \INMODEL{s\PRIME} which in
        turn is a constituent of the discourse situation d,
        if \INMODEL{a} is a constituent of \INMODEL{s};
        f(\INMODEL{s}) = undefined otherwise.\}.
\end{itemize}
For example, if the subset of {\DOMAIN{e}} in d consists of
the two atoms \INMODEL{j} and \INMODEL{b}, then \MGDVALUE{\TYPEDPAR{x}{e}} =
\{f$_1$,f$_2$,\ldots f$_i$,\ldots\}, where f$_1$, f$_2$ etc. are the functions
that may serve as the denotation of constants and variables---f$_1$ is the
function that maps each situation of which \INMODEL{j} is a constituent into
\INMODEL{j}, {f$_2$} is the function that maps each situation of which
\INMODEL{b} is a constituent into \INMODEL{b}--- and the other functions
represent all the possible denotations of objects that the parameter may be
resolved to. Note that the discourse situations plays here the role played by
the variable assignment in `free variable' theories of context dependence.

The grammar presented in the previous section can be straightforwardly extended
as follows to generate sentences such as \SENTENCE{It croaked}:

\begin{itemize}
  \item PRO  \RIGHTARROW it;  \TERM{\PAR{x}}
  \item NP \RIGHTARROW PRO;   \PSBRACKET{NP}{\PSBRACKET{PRO}{PRO\PRIME}}
\end{itemize}
The definition of the interpretation of logical forms  given above
already gives the correct results for these cases.

\subsubsection{Parameters and Discourse Interpretation}

Referential ambiguity gets `resolved' by \NEWTERM{\em anchoring} a parameter. A
parameter is \NEWTERM{anchored} if only one among the functions in its
denotation results in a consistent interpretation of the set of sentences in
which the parameter occurs; a parameter can be anchored by means of equality
statements of the form \INFIXEXP{=}{\PAR{x}}{a}, where \TERM{a} is not
parametric, or is already anchored: such equality statements make all but one
of the interpretations of the parameter inadmissible. Once a parameter is
anchored, it can be `replaced' by a term that denotes the one function among
those in the interpretation of the parameter that does not result in an
inconsistent interpretation, much as in the previous discussion of lexical
disambiguation, an H-type ambiguous predicate could be replaced by a
disambiguated version. So, the discourse interpretation principles formalizing
pronoun disambiguation involve a rewriting operation, just as the discourse
interpretation principles formalizing lexical disambiguation.

An apparent disadvantage of the present theory with respect to the `free
variable' theory of context dependence is that we can derive from the latter
that the value of referential expressions has to be fixed in order to get the
meaning of the sentence in which they occur. A conversation is infelicitous
unless the referents of all pronouns and definite descriptions have been
identified, the domain of quantification of all quantifiers has been
appropriately restricted, and so forth: so much so that listeners appear to be
ready to \NEWTERM{accomodate} new information (e.g., to introduce into the
discourse some otherwise unspecified antecedent for a pronoun) rather than
leave the interpretation unspecified \CITE{lewis:scorekeeping}.  But this
fact about referential expressions also follows if we treat context dependence
as a case of (H-type) semantic ambiguity; it is just a corollary of Pinkal's
precisification imperative, from which I derived the Condition on Discourse
Interpretation in section \SECREF{lexamb_section}. Accomodation procedures can
then be seen as a way of `precisifying' in lack of sufficient information.

\subsection{Syntactic Ambiguity}

The one case of ambiguity that requires extending the framework introduced here
considerably is syntactic ambiguity, as in \SENTENCE{They saw her
duck}. Furthermore, I haven't considered the problem of structural
disambiguation in any detail.  I refer the interested readers to
\cite{poesio:thesis,poesio:COGSCI95} for a sketchy discussion of how to encode
encoding syntactic ambiguity in an underspecified representation.\footnote{Most
systems making use of underspecified representations perform structural
disambiguation independently from the other forms of disambiguation
\cite{schubert&pelletier:82,allen:87,hobbs&shieber:87,alshawi:CLE-book}. There
is evidence, however, that structural disambiguation interacts at least with
reference interpretation \cite{crain&steedman:85,altmann&steedman:88} and a lot
of the recent work on statistical parsing relies on the hypothesis that lexical
interpretation affects parsing as well. Nothing in the proposal relies on
structural disambiguation occurring prior to the other stages of
disambiguation.}

\section{Discussion}

I have suggested that to develop a theory of discourse interpretation that is
consistent with what we know about the problem of ambiguity, we need to look
both at the grammar and at discourse interpretation. I proposed a theory of
grammar consistent with what I have called the {\em Underspecification
Hypothesis} and which is not based on the assumption that all natural language
expressions can be disambiguated; and a theory of discourse interpretation
according to which a perceived ambiguity occurs when defeasible interpretation
principles result in conflicting hypothesis. The interpretation process is
subject to two constraints: the {\em Anti-Random Hypothesis} (interpretations
are not generated at random) and the {\em Condition on Discourse
Interpretation}, derived from the {\em Precisification Imperative} (H-type
ambiguity has to be resolved). Although treatments of disambiguation based on
defeasible reasoning have been proposed elsewhere in the literature (e.g., in
\cite{alshawi:CLE-book}), I am not aware of any discussion of the
characteristics of this inferential process, the consequences of reasoning with
an underspecified representation, or the need for constraints on the inference
rules.

In the theory, semantic ambiguity is characterized model-theoretically in terms
of multiplicity of sense, whereas perceived ambiguity is characterized in terms
of inference. One may wonder if the distinction is really necessary; i.e., if
it is really the case that the meaning of natural language expressions can be
specified {\em a priori}. Two arguments in favor of a distinction are that it
provides for a clean distinction between the role of grammar and the role of
discourse interpretation; and that perceived ambiguity may also reflect
non-semantic distinctions, e.g., distinctions in speech act interpretation;
this question is not however totally resolved in the paper.

There are two obvious directions in which the present model needs to extended:
to provide a model of syntactic ambiguity, and to account for the effect of
incrementality in sentence processing. Preliminary work in this direction is
discussed in \cite{poesio:COGSCI95}.

An issue that deserves further inspection is whether the formal similarity
between the system used here to assign a denotation to indefinite sentences,
and the systems developed by Hamblin for dealing with questions
\cite{hamblin:questions} and by Rooth for its alternative semantics
\cite{rooth:thesis} has some significance. In particular, it would be
interesting to explore the consequences of using parameters as the translation
of focused elements.

\section*{Acknowledgments}

I owe the realization of the importance of the phenomenon of deliberate
ambiguity to my advisor Len Schubert and to Graeme Hirst. Special thanks to
Robin Cooper, Richard Crouch, Kees van Deemter, Howard Kurtzman, David Milward,
Manfred Pinkal for many discussions on the topic; and to two anonymous
reviewers whose suggestions went well beyond the call of duty. I am also
grateful to James Allen, Ariel Cohen, Tim Fernando, Janet Hitzeman, Peter
Lasersohn, Barbara Partee, Enric Vallduvi, Sandro Zucchi, and the audiences at
the University of Rochester, the University of Edinburgh, ICSI Berkeley,
Carnegie-Mellon University, University of Stuttgart, SRI Cambridge, Tilburg
University, and University of Saarbruecken.  All errors are of course
mine. This work was in part supported by the LRE Project 62-051 FraCaS.

\end{document}